\documentclass{amsart}
\usepackage[left=2.5cm, right=2.5cm]{geometry}
\usepackage{graphicx} 
\usepackage{amsmath}
\usepackage{color} 
\usepackage{amssymb}
\usepackage{amsthm}
\usepackage{accents}
\usepackage{float}
\usepackage{caption}
\usepackage[ruled,vlined,procnumbered]{algorithm2e}
\usepackage[english]{babel}
\usepackage{bbm}
\usepackage{chngcntr}
\usepackage{enumitem}
\usepackage{listings}
\usepackage{hyperref}
\usepackage{lmodern} 
\usepackage{mathtools}
\usepackage{setspace,graphicx,tikz,tabularx}
\mathtoolsset{showonlyrefs}
\newcommand{\R}{{\mathbb{R}}}

\def\rd#1{{\color{black}#1}}

\newtheorem{Theorem}{Theorem}[section]

\newtheorem{Proposition}{Proposition}[section]

\newtheorem{Remark}{Remark}[section]

\usepackage{fancyhdr}
\newcommand{\interior}[1]{%
  {\kern0pt#1}^{\mathrm{o}}%
}
\def\argmax{\mathop{\rm argmax}}
\def\1{{\bf 1}}

\title{P1-KAN: an effective Kolmogorov-Arnold network with application to hydraulic valley optimization}
\author{Xavier Warin  }
\address{Xavier Warin, EDF Lab Paris-Saclay and FiMe, Laboratoire de Finance des March\'es de l'Energie, 91120 Palaiseau, France}
 \email{xavier.warin@edf.fr}

\begin{document}
\begin{abstract}
We propose a new Kolmogorov-Arnold network (KAN), the P1-KAN, to approximate irregular functions in high dimensions.
We present universal approximation theorems for its various forms. We also derive approximation errors for the special case in which the Kolmogorov-Arnold representation functions are regular.
Based on our analysis of simple regression errors, we demonstrate that the P1-KAN outperforms multilayer perceptrons in terms of accuracy and convergence speed.
Additionally, we compare the P1-KAN with several other proposed KAN networks and demonstrate that the P1-KAN \rd{is the most effective} for irregular functions  and achieves accuracy similar to that of the original spline-based KAN network for smooth functions.
Finally, we compare KAN networks when optimizing a French hydraulic valley. In this industrial application, the P1-KAN \rd{optimizes better than other networks}. It also \rd{optimizes better} than classical deterministic tools based on dynamic programming solvers used by practitioners.
\end{abstract}
\maketitle
\section{Introduction}
Kolmogorov-Arnold Networks \rd{(KANs)} \cite{liu2024kan}, based on the Arnold-Kolmogorov representation theorem, have recently been proposed as an alternative to multilayer perceptrons \rd{(MLPs)} for approximating functions in high dimensions.
Arnold and Kolmogorov showed long ago \cite{kolmogorov1961representation} that a multivariate continuous  function $f$ on a compact set can be written as a finite composition of the sum of continuous functions of a single variable. More precisely, if $f$ is continuous on $[0,1]^n$, then
\begin{align}
    f(x)= \sum_{i=1}^{2n+1} \psi_i\left(\sum_{j=1}^n \Phi_{i,j}(x_j)\right),
    \label{eq:kom}
\end{align}
where $\Phi_{i,j}: [0,1] \longrightarrow \mathbb{R}$ and $\psi_i : \mathbb{R} \longrightarrow \mathbb{R}$.\\ 
As the 1D functions can be very irregular, it has been shown that they may not be learnable in practice \cite{girosi1989representation,poggio2020theoretical}.
To overcome this limitation, \cite{liu2024kan} propose to extend this representation.
First, they propose not to stick to $2n+1$ terms in the outer sum in \eqref{eq:kom} and to define a KAN $l^{th}$ layer as an operator $\psi^l_{m,q}$ from $[0,1]^{m}$ to $\mathbb{R}^{q}$:
\begin{align}
  (\psi^l_{m,q}(x))_k=  \sum_{j=1}^m \Phi_{l,k,j}(x_j), \text{ for } k=1, \ldots, q.
  \label{eq:PK1L}
\end{align}
Second, by stacking the layers, i.e., composing the operator $\psi^l$, they define the KAN operator from $[0,1]^m$ to $\mathbb{R}^d$:
\begin{align}
    K(x)= (\psi^L_{n_{L-1},d} \circ \psi^{L-1}_{n_{L-2}, n_{L-1}} \circ \ldots \circ \psi^1_{n_0,n_1} \circ \psi^0_{m,n_0}) (x)
    \label{eq:KAN}
\end{align}
Since all $\psi$ functions are one-dimensional, many classical methods are available to propose an easy-to-implement approximation.
In their proposed implementation \rd{(that we denote Spline-KAN)}, \cite{liu2024kan} use B-splines (see for example \cite{chaudhuri2021b}) associated with the SILU activation function to approximate the $\psi$ function: the spline coefficients and the multiplicative coefficient of the SILU function are learned using a classical stochastic gradient algorithm as done with MLPs.\\
This network has been rapidly tested, replacing MLPs in transformers \cite{yang2024kolmogorov} for example, and in various fields: the medical sector in \cite{knottenbelt2025coxKAN}, vision \cite{cheon2024demonstrating,li2025u}, time series \cite{vaca2024kolmogorov},\cite{xu2024kan}, \cite{xu2024kolmogorov}.
Strengths and weaknesses of this approach compared to MLPs are discussed in \cite{yu2024KAN} and, depending on its use, its superiority to MLPs is not always obvious \cite{shen2025reduced,le2024exploring}.\\
Following this first article, different evolutions of this architecture are proposed to address different problems: \cite{genet2024tKAN} proposes an evolution of the algorithm to replace LSTM in time series, \cite{xu2024fourierKAN} for Graph Collaborative Filtering, \cite{bodner2024convolutional} for convolutional networks, \cite{abueidda2025deepokan} in mechanics.\\
The original spline-based algorithm has several drawbacks.
The first disadvantage of this approach is that the spline approximation is expensive, at least in the original algorithm proposed.
The second is that the output of a layer may not be in the grid initially chosen for the following layer. Finally, since the Kolmogorov representation theorem may involve a very irregular function, one may wonder whether it is interesting to use a rather high-order approximation such as a spline.\\
To address the first point, many other approximations based on classical numerical analysis have been proposed using: wavelets \cite{bozorgasl2024wav}, radial basis \cite{li2024kolmogorov,ta2024bsrbf} which reduces the computation time by 3, Chebyshev polynomials \cite{ss2024chebyshev} and many others. An interesting representation that leads to a very effective layer is the ReLU-KAN \cite{qiu2025relu} \cite{so2025higher}, which is based only on the ReLU function, matrix addition and multiplication, and divides the computation time by 20. \\
To address the second point, some use a sigmoid activation function \cite{ss2024chebyshev} to obtain an output in the range $[0,1]$, but this approach is clearly very ineffective. Others attempt to adapt the support of the basis functions by trying to learn them \cite{liu2024kan}, but this adaptation often fails.\\
\rd{On the theoretical point on of view, no universal approximation theorem with the sup norm for $C^0(K)$ functions is available for a given compact $K$.
A recent article \cite{kratsios2025kolmogorov} gives convergence guarantees  for the Res-KANs (a modified version of the Spline-KAN)  on the Bezov space $B^\alpha_{p,q}$ but as $B^0_{\infty,\infty}$ is not included in the framework, it does not cover $C^0(K)$.}\\
\rd{All the methods proposed in the literature face two main issues:
\begin{itemize}
    \item None of them are theoretically justified by a universal approximation 
    theorem in the supremum norm on compact sets, and their efficiency remains 
    poorly understood.
    \item Since they rely on bases of regular functions, they are expected to be 
    less effective when approximating irregular target functions, so functions $C^0$ and not $C^1$ for example.
\end{itemize}
}
Concerned by the possibility of KANs to approximate high-dimensional functions, especially for stochastic optimization purposes in \cite{germain2021neural},\cite{warin2023reservoir} \rd{with  a theoretical background}, we have developed the P1-KAN network, borrowing some interesting features from the ReLU-KAN, but clearly defining the support of the layer function and avoiding the network adaptation proposed in \cite{liu2024kan}. The name P1-KAN is related to the classical finite element method, which uses P1 hat functions for interpolation \cite{quarteroni2009numerical}. 
\rd{The P1-KAN is fundamentally different from the ReLU-KAN, which uses smooth basis functions constructed solely from ReLU functions in order to reduce computational cost. The resulting approximation by the ReLU-KAN is not piecewise linear.
The Fast-KAN network \cite{li2024kolmogorov}, also used in this work, relies on radial basis functions and also aims to reduce the computational cost compared to the Spline-KAN. }
The proposed P1-KAN network concatenates layers coherently\rd{, so differently from existing KANs}, enabling us to provide error bounds under the assumption that the functions in expansion \eqref{eq:kom} are Lipschitz, which generally does not apply to outer functions.

 We provide universal approximation theorems to address the more general case.  These are the first  universal approximation theorems \rd{ for the sup  normal for continuous function on a compact} to be provided for KANs.
 
In the second part, we compare it with MLPs, Spline-KAN, Radial basis KAN, and ReLU-KAN on function approximation using either smooth or very irregular functions in different dimensions.
Finally, we test the Spline-KAN and the P1-KAN to optimize a French hydraulic valley, comparing them with classical perceptrons.

\section{The P1-KAN Networks}
\label{sec:P1Kanlayer}
We will first explain the main features of the method, and then go into detail about the algorithm and why P1-KAN is different from other KAN networks.
\subsection{The P1-KAN Layers}
The first possible layer uses a regular grid, while the second one uses a grid adapting to the data.
\subsubsection{The P1-KAN Layer with a Regular Lattice}
\label{sec:P1KAnLayerNodAdapt}
\begin{figure}[H]
    \centering
    \includegraphics[width=0.25\linewidth]{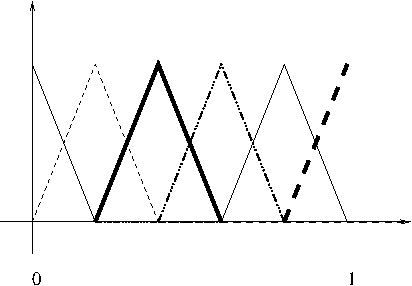}
    \caption{Uniform $P_1$ basis functions on $[0,1]$ with $P=5$.}
    \label{fig:P1}
\end{figure}

\hfill\\
We assume that layer $l$ is an operator  with support $G^l=(\underline{x}^l, \bar{x}^l) \in \mathbb{R}^{d_0 \times 2}$ and with values in $\mathbb{R}^{d_1}$. 
As for the classical KAN layers, a number of meshes per direction $P$ (assumed constant per layer to simplify the notations) are used to discretize $[\underline{x}_1^l, \bar{x}_1^l] \times \ldots \times [\underline{x}_{d_0}^l, \bar{x}_{d_0}^l]$, giving the mesh vertices $(\hat{x}_{j,p}^l)_{1 \le p \le P-1}$ in $]\underline{x}_j^l, \bar{x}_j^l[$ for $j=1, \ldots,d_0$. 
Set $\hat{x}_{j,0}^l= \underline{x}_j^l$, $\hat{x}_{j,P}^l=\bar{x}_j^l$, 
the function $\Phi_{l,k,j}$ in \eqref{eq:PK1L} is defined using a P1 finite element method (see figure \ref{fig:P1}): for $x \in [\underline{x}_j^l, \bar{x}_j^l]$,
\begin{align}
\Phi_{l,k,j}(x) = \sum_{p=0}^{P} a^{l,k,j}_p \Psi^{l,j}_p(x)
   \label{eq:P1}
\end{align}
where $(a^{l,k,j}_p)_{p=0,\ldots,P}$ are  trainable variables and $(\Psi^{l,j}_p)_{p=0,\ldots,P}$ is the basis of the shape function $\Psi^{l,j}_p$ with compact support $[\hat{x}_{j,p-1}^l, \hat{x}_{j,p+1}^l]$ for $p=1, \ldots, P-1$ and defined as:
\begin{align}
    \Psi^{l,j}_p(x) = \left \{
    \begin{array}{cc}
    \frac{x - \hat{x}^l_{j,p-1}}{\hat{x}^l_{j,p} - \hat{x}^l_{j,p-1}} &  x \in [ \hat{x}^l_{j,p-1} , \hat{x}^l_{j,p}] \\
    \frac{\hat{x}^l_{j,p+1} - x}{\hat{x}^l_{j,p+1} - \hat{x}^l_{j,p}} &  x \in [ \hat{x}^l_{j,p} , \hat{x}^l_{j,p+1}] \\    
    \end{array}
    \right.  \nonumber \\
    \text{ such that }  \Psi^{l,j}_p(\hat{x}^l_{j,q}) = \delta_{q,p},
    \label{eq:shape}
\end{align}
for $p=1, \ldots, P-1$.
Similarly, $\Psi^{l,j}_0$ (or $\Psi^{l,j}_{P}$) is defined as a continuous  piecewise linear function with support $[\underline{x}^l_j , \hat{x}^l_{j,1}]$ (or $[\hat{x}^l_{j,P-1}, \bar{x}^l_j]$) and such that $\Psi^{l,j}_{0}(\underline{x}^l_j)=1$ (or $\Psi^{l,j}_{P}(\bar{x}^l_j)=1$).
Unlike other networks, the P1-KAN layer, which is theoretically described as an operator from $\mathbb{R}^{d_0}$ to $\mathbb{R}^{d_1}$ by the equation
\eqref{eq:PK1L}, takes as input not only a sample $x \in \mathbb{R}^{d_0}$ but also the description of the support $(\underline{x}^l_j, \bar{x}^l_j)_{j=1,d_0}$.\\
In this version of the layer, the grid is uniform on $[\underline{x}^l_j, \bar{x}^l_j]$, and the vertices are generated for each direction $j$ for $0 \le p \le P$ by
\begin{align}
    \hat{x}^l_{j,p} = \underline{x}^l_j +  p \frac{\bar{x}^l_j - \underline{x}^l_j}{P}.
    \label{eq:regGrid}
\end{align}
The operator estimating the value function associated to the layer on the hypercube $G^l$ is given by:
\begin{align}
  \hat \kappa^{l,P}_{d_0,d_1}(x, G^l)_k = \sum_{j=1}^{d_0} \sum_{p=0}^{P} a^{l,k,j}_p \Psi^{l,j}_{p}(x_{j}), \text{ for }  k=1, \ldots, d_1.
  \label{eq:PK1LL}
\end{align}
The tensor $A^l_{d_0,d_1}=(a^{l,k,j}_p)_{0 \le p \le P, 1 \le j \le d_0, 1 \le k \le d_1}$ are the trainable variables of the network for layer $l$.\\
As output, the layer returns the values of $\hat \kappa^{l,P}_{d_0,d_1}(x, G^l)$ in $\mathbb{R}^{d_1}$
and the lattice $G^{l+1} = [ \underline{G}^{l+1}, \bar{G}^{l+1}]$ obtained from the possible $ \hat \kappa^{l,P}_{d_0,d_1}(x, G^l)$ values.
Due to the use of the $P_1$ finite element approximation, this output lattice is exactly obtained from the $A^l_{d_0,d_1}$ tensor by:
\begin{align*}
    \underline{G}^{l+1}_k = &\sum_{j=1}^{d_0} \min_{0 \le p \le P} a^{l,k,j}_p \\
    \bar{G}^{l+1}_k = & \sum_{j=1}^{d_0} \max_{0 \le p \le P} a^{l,k,j}_p,
\end{align*}
for $1 \le k \le d_1$.\\
Then the global layer as  an operator from $\R^{d_0} \times  \R^{d_0 \times 2} $ to $\R^{d_1} \times \R^{d_1 \times 2}$ is defined by:
\begin{align}
  \kappa^{l,P}_{d_0,d_1}(x, G^l) = (  \hat \kappa^{l,P}_{d_0,d_1}(x, G^l), G^{l+1}).
   \label{eq:PK1LLGlob}
\end{align}

\subsubsection{The P1-KAN Layer with an Adapting Lattice}
\hfill\\
In this version of the layer, the vertices in $]\underline{x}^l_j, \bar{x}^l_j[$ are initially generated randomly uniformly and their values are trained to adapt to the data. 
This is done by defining for $1 \le p < P$ increasing values in $]0,1[$ by $\frac{\sum_{k=1}^p e_k}{\sum_{k=1}^P e_k}$ where $e_1, \ldots, e_P$ are positive random variables. This set of values is then used to define the grid points in $]\underline{x}^l_j, \bar{x}^l_j[$ by an affine  transformation.\\

In detail, the vertices in $]\underline{x}^l_j, \bar{x}^l_j[$ are generated for each direction $j$ for $1 \le p < P$ by
\begin{align}
    \hat{x}^l_{j,p} = \underline{x}^l_j + (\bar{x}^l_j - \underline{x}^l_j) \frac{\sum_{k=1}^p \exp{(y^{l,k,j})}}{\sum_{k=1}^{P} \exp{(y^{l,k,j})}},
    \label{eq:adaptGrid}
\end{align}
where the matrix $Y^l_{d_0}= (y^{l,p,j})_{1 \le p \le P, 1 \le j \le d_0}$ has elements in $\mathbb{R}$.
The tensor $A^l_{d_0,d_1}=(a^{l,k,j}_p)_{0 \le p \le P, 1 \le j \le d_0, 1 \le k \le d_1}$ and $Y^l_{d_0}$ are the trainable variables of the layer.
\rd{The trainable variables $(y^{l,p,j})_{1 \le p \le P, 1 \le j \le d_0}$ are initialized at random. An alternative approach would be to initialize them such that the $(\hat{x}^l_{j,p})_{0 \le p \le P}$ values are initially uniformly distributed.}
\subsubsection{Importance of Adaptation}
\label{sec:importsamp}
\begin{figure}[H]
\begin{minipage}[b]{0.32\linewidth}
  \centering
 \includegraphics[width=\textwidth]{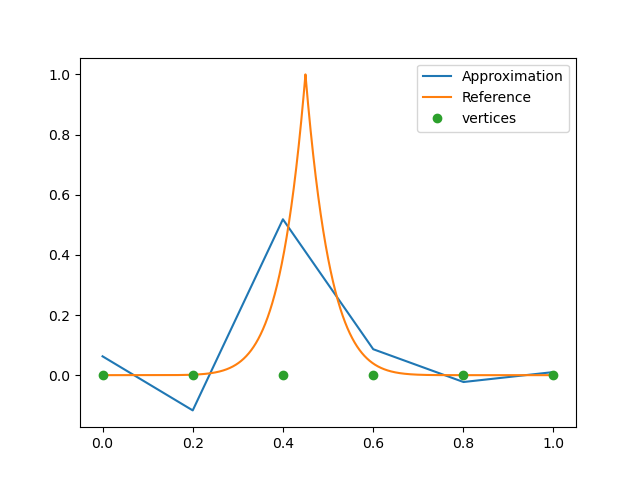}
 \caption*{\tiny $P=5$ No adaptation}
 \end{minipage}
 \begin{minipage}[b]{0.32\linewidth}
  \centering
 \includegraphics[width=\textwidth]{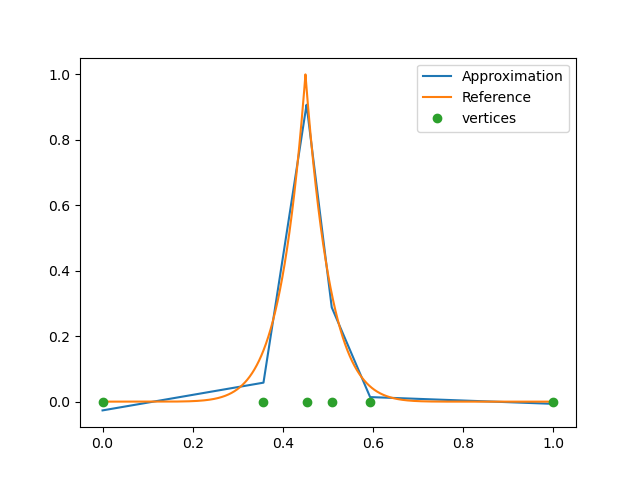}
 \caption*{\tiny $P=5$ Adaptation}
 \end{minipage}
 \begin{minipage}[b]{0.32\linewidth}
  \centering
 \includegraphics[width=\textwidth]{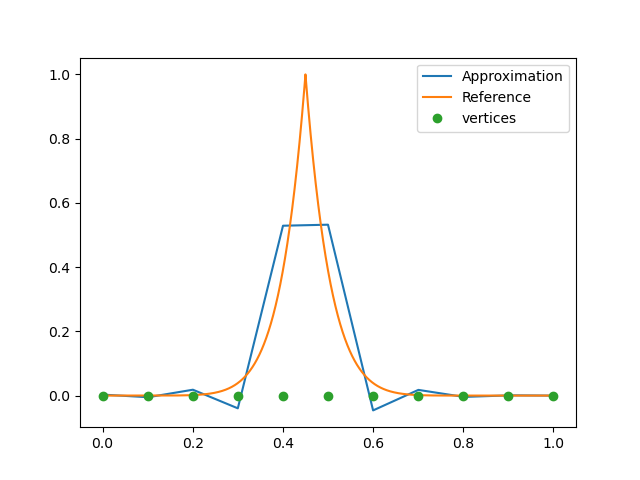}
 \caption*{\tiny $P=10$ No adaptation}
 \end{minipage}
 \begin{minipage}[b]{0.32\linewidth}
  \centering
 \includegraphics[width=\textwidth]{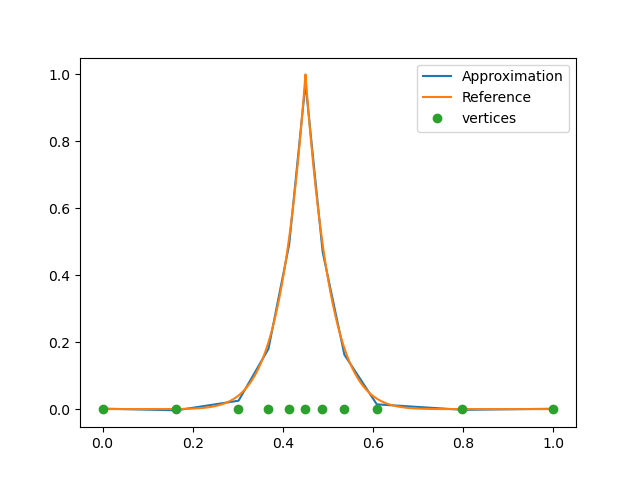}
 \caption*{\tiny $P=10$ Adaptation}
 \end{minipage}
 \begin{minipage}[b]{0.32\linewidth}
  \centering
 \includegraphics[width=\textwidth]{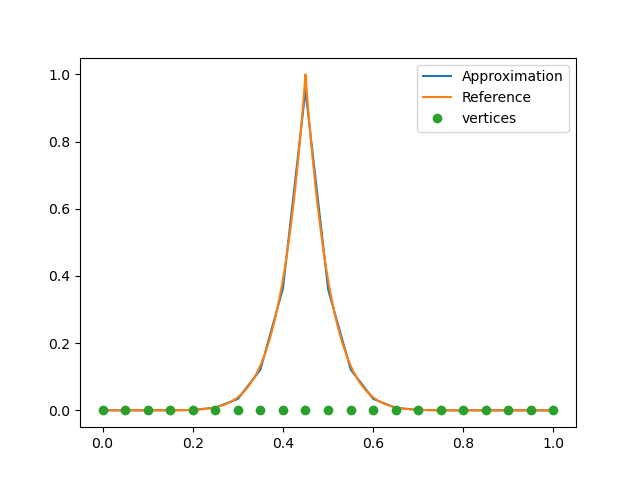}
 \caption*{\tiny $P=20$ No adaptation}
 \end{minipage}
 \begin{minipage}[b]{0.32\linewidth}
  \centering
 \includegraphics[width=\textwidth]{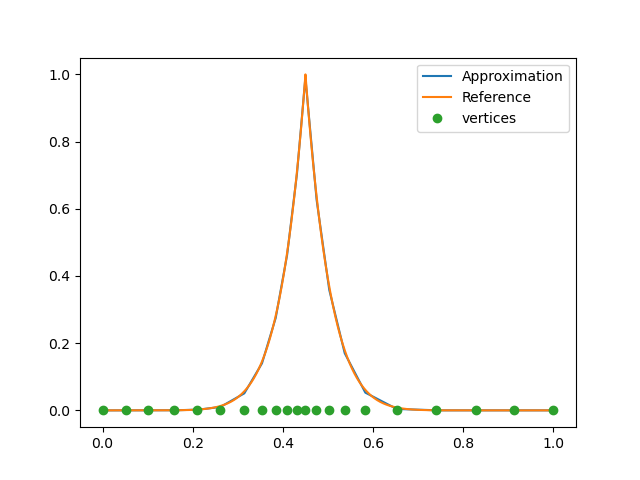}
 \caption*{\tiny $P=20$ Adaptation}
 \end{minipage}
 \caption{Reference function and approximation with or without adaptation. Vertices on the x-axis : \rd{ adaptation is crucial for a good approximation.} \label{fig:adapt1D}}
\end{figure}

To illustrate the importance of adaptation, we minimize the mean squared error in one dimension of our P1-KAN layer approximation, denoted $f_{P}(x)$, with respect to the one-dimensional function $f(x)= x^8 1_{x < 0.45} +(0.9-x)^8 1_{x \ge 0.45}$ on $[0,1]$ for different values of $P$.
In Figure \ref{fig:adapt1D}, we plot $f_P$, $f$ and the vertices of the lattice with and without adaptation.
As expected, adaptation increases the number of points in areas where the gradient is the highest.
The mean squared error obtained after optimization is given in Table \ref{tab:adapt1D}. The cost of a calculation using adaptation is less than twice the cost of a calculation without adaptation.
\begin{table}[H]
    \centering
    \begin{tabular}{|c|c|c|} \hline
        $P$ &  Error without adaptation &  Error with adaptation \\ \hline
        5 &   0.1545 & 0.00492 \\
        10 & 0.0737 &  3.55E-4 \\
        20 & 4.7E-4 &  5.22E-5 \\ \hline
    \end{tabular}
    \caption{Mean squared error \rd{showing that adaptation is necessary for low values of $P$}.}
    \label{tab:adapt1D}
\end{table}
\subsection{The Global P1-KAN Network}
As shown in the previous section, the P1-KAN layer inputs $x$ values and a hypercube, and it outputs the values obtained by the operator and a hypercube. Therefore, it is natural to stack the layers without using any grid adaptation or sigmoid function to send the output of the layer back to a known bounded domain. The P1-KAN network takes as the initial hypercube used for the first layer the hypercube corresponding to the bounded domain where we want to approximate the unknown function.\\
Supposing that the number of neurons is equal to $N$, the number of inside layers is $L$, the global operator for $x \in \mathbb{R}^n$ is
\begin{align}
    K(x)= \hat \kappa^{L,P}_{N,1} \circ  \kappa^{L-1,P}_{N,N}  \circ \ldots  \circ  \kappa^{0,P}_{n,N}(x, [0,1]^n),
    \label{eq:stackP1L}
\end{align}
where the $\kappa^{l,P}$ are defined in \eqref{eq:PK1LLGlob} ,
 $\hat  \kappa^{L,P}$ in \eqref{eq:PK1LL} and  the parameters to optimize are
$ \theta ={\mathcal A} := A^0_{n,N} \cup A^1_{N,N} \cup \ldots \cup A^{L}_{N,1}$ for the P1-KAN network without adaptation, and $\theta ={\mathcal A}  \cup Y^0_{n} \cup Y^1_{N} \cup \ldots \cup Y^{L}_{N}$ for the P1-KAN with adaptation.\\
The approximation space spanned by the P1-KAN parametrized by $\theta$ for a number of meshes depending on the layer is then for $P =(P_0,\ldots, P_L)$:
\begin{align}
  {\mathcal N}^{L,N,P}_{n} = \{ f_{\theta}(x) = \hat \kappa^{L,P_L}_{N,1} \circ \kappa^{L-1,P_{L-1}}_{N,N} \circ \dots \circ \kappa^{0,P_0}_{n,N}(x, [0,1]^n)  \mbox{with }  \theta \in \mathbb{R}^Q\},
  \label{eq:spaceSpawn}
\end{align} 
where $Q= nN (P_0+1) +  \sum_{k=1}^{L-1} (P_k+1) N^2+  (P_L+1) N $ for the P1-KAN without adaptation and $Q=  n(N (P_0+1)+ (P_0-1)) +  \sum_{k=1}^{L-1} [(P_k+1) N^2 + (P_k-1) N]+  2 P_L N $ with adaptation.
\\
\rd{ An implementation in Pytorch is available at \href{https://gitlab.com/shared5530016/p1-kan}{https://gitlab.com/shared5530016/p1-kan}}.\\

\subsection{Convergence theorems}
The Kolmogorov-Arnold theorem has been refined in several ways:
\begin{itemize}
\item The functions  can be chosen from the class of $\alpha$- Hölder functions ($0 < \alpha < 1$) \cite{lorentz1966approximation} or even Lipschitz functions \cite{fridman1967improvement}.
\item The  $\Phi_{i,j}$  functions cannot be constructed in the class of  $C^1$ functions \cite{vitushkin1964proof}, \cite{henkin1964linear}.
\item By the construction technique used, the  functions  $\psi_i$ are no more than continuous.
\end{itemize}
Note that if we accept an approximation of $f$ that is supposed to be $\alpha$-Hölder ($0 < \alpha < 1$), \eqref{eq:kom} can be modified using $N$ term on the outer summation leading to the construction of an approximation $f_N$ with $\Phi_{i,j}$ $C^2$ converging to $f$ in the  sup norm as $N$ increases \cite{song2025explicit} (see also \cite{demb2021note} for another  construction approximating the function in the sup norm).\\
Even though the constructions proposed in the literature use only continuous $\psi_i$ functions, we can consider the space spanned by \eqref{eq:kom} using Lipschitz $\Phi_{i,j}$  and $\psi_i$ functions.\\
We begin by providing convergence estimations for the aforementioned space using classical methods in numerical analysis. We emphasize that this is not the general case, even when we suppose that $f$ is Lipschitz.  The general case, which yields only a universal approximation theorem, is addressed in the final sub section.

\subsubsection{Supposing regularity of the outer functions}
Let us define the interpolation operator $\Pi_P^{[a,b]}$ of a function $f$ defined on $[a,b] \subset \mathbb{R}$, with a step $h= \frac{b-a}{P}$ by
\begin{align}
    \Pi_P^{[a,b]} (f)(x)=  \sum_{i=0}^P f( a + ih) \Psi_{i}(x),
\end{align}
where $\Psi_{0}(x)= \max( 1- \frac{x -a}{h},0 )$, $\Psi_{P}(x)= \max( \frac{x-b+h}{h},0 )$, and for $i=1, \ldots , P-1$,
\begin{align}
    \Psi_{i}(x) = \left \{
    \begin{array}{cc}
    \frac{x - (a + (i-1)h)}{h} &  x \in [ a + (i-1)h , a + ih] \\
    \frac{ (a + (i+1)h) - x}{h} &  x \in [ a + ih , a + (i+1)h]. \\   
    \end{array}
    \right.
\end{align}
For a set of Lipschitz functions $f_i :  [0,1] \longrightarrow \mathbb{R}$ for $i=1,\ldots, n$, we denote  
\begin{align*}
 \mu( \{f_i\}_{i=1,n}) := &[ \underline{\mu}( \{f_i\}_{i=1,n}), \bar{\mu}( \{f_i\}_{i=1,n})] \\
 :=& [ \min_{x \in   [0,1]^n}  \sum_{i=1}^n f_i(x_i), \max_{x \in [0,1]^n }  \sum_{i=1}^n f_i(x_i) ],
\end{align*}
which is well defined by continuity on a compact.
 
\begin{Proposition}
\label{prop:interp}
    Supposing that the $ (\psi_i)_{i=1, \ldots, 2n+1}$ and the $(\Phi_{i,j})_{i=1,\ldots,2n+1 , j=1, \ldots,n}$ in the Kolmogorov-Arnold expansion \eqref{eq:kom} are $K$-Lipschitz. For $h>0$ given, let us define $$P = \lfloor \frac{1}{h} \rfloor +1, $$ $$ \hat{P}_i =  \lfloor \frac{\bar{\mu}( \{\Pi^{[0,1]}_P (\Phi_{i,j})\}_{j=1,n}) -\underline{\mu}( \{\Pi^{[0,1]}_P (\Phi_{i,j})\}_{j=1,n})}{h}\rfloor +1, $$ for  $i=1, \ldots 2n+1$, then there exists a constant $C$ such that
    \begin{align}
    \sup_{x \in [0,1]^n} | f(x) - \sum_{i=1}^{2n+1}  \Pi^{\mu(\{ \Pi^{[0,1]}_P(\Phi_{i,j})\}_{j=1,\ldots,n})}_{\hat{P}_i} (\psi_i)(\sum_{j=1}^n  \Pi^{[0,1]}_P(\Phi_{i,j})(x_j)) | \le C h.
    \end{align}
\end{Proposition}
\begin{proof}
    By classical estimates on $P_1$ interpolation on Lipschitz functions, there exists $D$ such that
    \begin{align}
    \label{eq:Interp}
         \sup_{x \in [0,1]} |\Pi^{[0,1]}_P (\Phi_{i,j})(x) -\Phi_{i,j}(x) | \le D h.
    \end{align}
    Then
    \begin{align*}
      A =& | f(x) - \sum_{i=1}^{2n+1}  \Pi^{\mu(\{ \Pi^{[0,1]}_P(\Phi_{i,j})\}_{j=1,\ldots,n})}_{P_i} (\psi_i)(\sum_{j=1}^n  \Pi^{[0,1]}_P (\Phi_{i,j})(x_j)) |  \\
      \le  & | f(x)-  \sum_{i=1}^{2n+1}  \psi_i(\sum_{j=1}^n  \Pi^{[0,1]}_P (\Phi_{i,j})(x_j)) |
      + | \sum_{i=1}^{2n+1}  (\psi_i-\Pi^{\mu(\{ \Pi^{[0,1]}_P(\Phi_{i,j})\}_{j=1,\ldots,n})}_{P_i} (\psi_i)) (\sum_{j=1}^n  \Pi^{[0,1]}_P(\Phi_{i,j})(x_j)) |\\
      \le & K \sum_{i=1}^{2n+1} |\sum_{j=1}^n  \Pi^{[0,1]}_P (\Phi_{i,j})(x_j) - \Phi_{i,j}(x_j)| + (2n+1) D h \\
      \le &  (2n+1)D(Kn+1)h.
    \end{align*}
    using \eqref{eq:Interp}.
\end{proof}

\begin{Proposition}
\label{prop:conv1}
    Under the conditions and definitions  of Proposition \ref{prop:interp},  the P1-KAN without adaptation defined by \eqref{eq:stackP1L} with $L \ge 1$ and $N \ge 2n+1$ satisfies:
\begin{align}
   \min_{ g \in { \mathcal N}^{L,N,\bar P}_{d}}  \sup_{x \in [0,1]^d} |f(x)- g(x)| \le C h.
\end{align}
where $\bar P= (P,  P_1, \ldots , P_L)$,  $P_1=\sup_{i =1, \ldots, 2n+1}  \hat P_i$, and  $P_2, \ldots P_L$ chosen  strictly positive. 

\end{Proposition}
\begin{proof}
As the identity and the zero functions can be generated by the functions $\Psi^{l,j}_{p}$ in \eqref{eq:PK1L},
\begin{align*}
   \min_{ g \in { \mathcal N}^{L,N,\bar P}_{d}}  \sup_{x \in [0,1]^d} |f(x)- g(x)| \le&  \min_{ g \in { \mathcal N}^{1,2n+1,(P,P_1)}_{d}}  \sup_{x \in [0,1]^d} |f(x)- g(x)| \\
   \le & C \sup_{x \in [0,1]^d} | f(x) - \sum_{i=1}^{2n+1}  \Pi^{\mu(\{\Pi^{[0,1]}_P(\Phi_{i,j})\}_{1,\ldots,d})}_{\hat{P}_i} (\psi_i)(\sum_{j=1}^n  \Pi^{[0,1]}_P(\Phi_{i,j})(x_j)) | \\
   \le &C h ,
\end{align*}
by Proposition \ref{prop:interp}.
\end{proof}
\begin{Remark}
As the non-adaptation case is a special case of the adaptation case, Proposition \eqref{prop:conv1} still holds with adaptation.
\end{Remark}
\rd{
\begin{Remark}
Assuming that the Kolmogorov--Arnold decomposition involves more regular 
functions, the error estimate could be of order $O(h^{2})$. A similar error 
expression can be derived for the Spline-KAN if the layers are concatenated coherently, which would yield a higher 
convergence rate (depending on the spline order).
When the functions involved are only Lipschitz continuous, the use of splines 
should not improve the results.
\end{Remark}
}

\subsubsection{Universal approximation theorems}
As stated in the introduction, the previous propositions only apply when the Kolmogorov–Arnold outer functions are Lipschitz, {\bf which is not generally the case}. Therefore, a universal approximation theorem is required to ensure convergence of the network :
\begin{Theorem}
    The space spanned by ${\mathcal N}^{L,N,P \1_{L+1}}_n$ letting $N$ vary for $L \ge 1$, $P>1$ is dense in $C^0([0,1]^n)$ with the sup norm when adaptation is used.
    \label{theo1}
\end{Theorem}
\begin{proof}
We first prove the result for $P=2$, $L=1$.
Using \cite{hornik1990universal}, for all functions $f \in C^0([0,1]^n)$, there exist $N$, $A \in \R^N$, $B \in \R^{N\times n}$, $C \in \R^N$ such that:
\begin{align}
\sup_{x \in [0,1]^n} |f(x) -  \sum_{i=1}^N A_i \max( B_i x + C_i, 0)| \le \epsilon,
\end{align}
where $B_i$ is the row vector defined by $(B_{i,j})_{j=1,n}$.
It is clear that one can find $\hat \kappa^{0,2}_{n,N}$ defined by 
\eqref{eq:PK1LL} such that
$ \hat \kappa^{0,2}_{n,N}(x,[0,1]^n)_i =  B_i x + C_i$ for $i=1, \ldots, N$.
For each $i$ on $G^1_i=[ \inf_{x \in [0,1]^n}  B_i x + C_i, \sup_{x \in [0,1]^n}  B_i x + C_i]$, $g_i(x)= A_i \text{ReLU}(x)$ is either linear, or $0 \in \interior {G^1_i}$ so 
$g_i(x)$ can be generated by a $\hat \kappa^{1,2}_{N,1}(x,G^1)_i$ using adaptation so that $\hat x^{1}_{i,1}=0$ when $0 \in \interior {G^1_i}$.\\
The proof can easily be extended to $P>2$ using the fact that a linear-by-part function, when discretized with $P=2$ meshes, can also be represented with $P>2$ meshes due to adaptation.
Finally, the result is extended to $L > 1$ layers using the fact that the identity and zero functions are generated by the P1 hat functions and can be used to add unnecessary layers.
\end{proof}
When adaptation is not used, the ReLU function cannot directly be generated as $0$ is generally not in the lattice. The ReLU function is therefore uniformly approximated by piecewise linear functions by increasing $P$, which gives the second theorem:
\begin{Theorem}
 The space spanned by ${\mathcal N}^{L,N,P \1_{L+1}}_n$ letting $N$ and $P$ vary for $L \ge 1$ is dense in $C^0([0,1]^n)$ with the sup norm when adaptation is not used.    
\end{Theorem}

\begin{Remark}
    Theoretically, $P=2$ is sufficient to replicate the classical feedforward behavior \rd{when adaptation is used}.
\end{Remark}

\section{Numerical Results for Function Approximation}
In this section, we compare the classical feedforward network with Spline-KAN  \cite{liu2024kan}, Fast-KAN \cite{li2024kolmogorov}, ReLU-KAN \cite{qiu2025relu}, and P1-KAN for the approximation of  two types of functions defined on $[0,1]^n$.
\begin{figure}[H]
\centering
  \begin{minipage}[b]{0.49\linewidth}
  \centering
 \includegraphics[width=\textwidth]{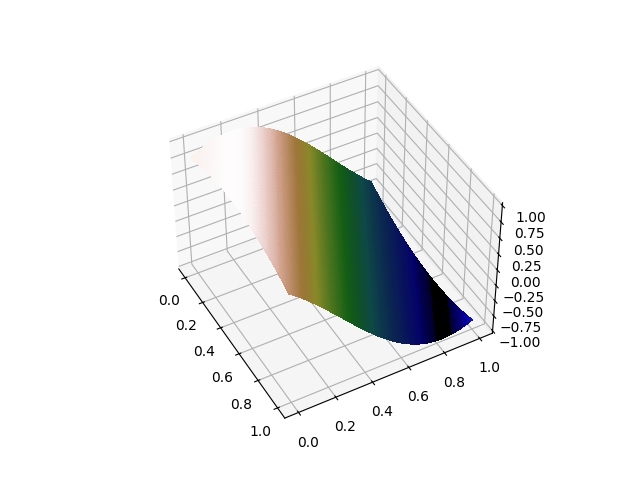}
 \caption{\label{fig:A} Function A in 2D.}
 \end{minipage}
   \begin{minipage}[b]{0.49\linewidth}
  \centering
 \includegraphics[width=\textwidth]{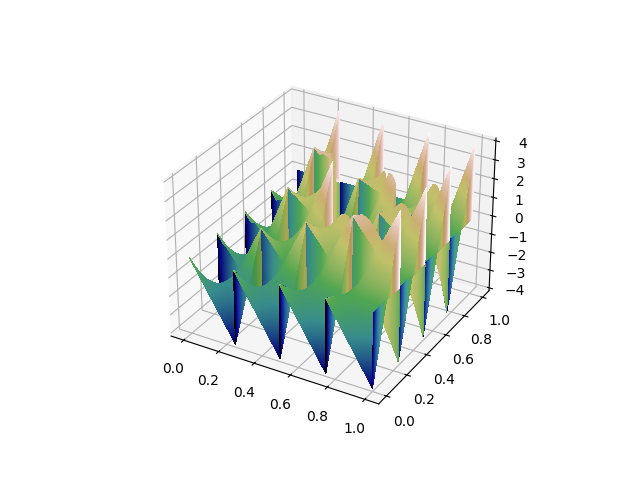}
 \caption{\label{fig:B} Function B in 2D.}
 \end{minipage}
\end{figure}
\begin{enumerate}
    \item[A] The first function is smooth but very fast oscillating with increasing dimension and is defined for $x \in [0,1]^n$ by:
    \begin{align}
        f(x)=  \cos\left( \sum_{i=1}^n  i y_i \right),
    \end{align}
    where $y= 0.5 +\frac{2x-1}{\sqrt{n}}$.
    The function in 2D is shown in Figure \ref{fig:A}.
     It becomes increasingly difficult to approximate as the dimension grows, and classical feedforward generally fails in dimensions above 12.
    \item[B] The second function is very irregular and is given as
     \begin{align}
        f(x)=  n \left(\prod_{i=1}^n  y_i + 2 \left( 4 \prod_{i=1}^n x_i - \lfloor 4 \prod_{i=1}^n x_i \rfloor \right)-1\right),
    \end{align} 
    where $y := 2 \left( 4x - \lfloor 4x \rfloor \right)-1$ and for $x \in \mathbb{R}^n$, $\lfloor x \rfloor := (\lfloor x_i \rfloor)_{i=1,n}$.
    The function is shown in 2D in Figure \ref{fig:B}.
    Even if the function is discontinuous, it is interesting to see  how the networks behave on this extreme case.
    
\end{enumerate}
To approximate a function $f$ with a neural network $\kappa$, we use the classical quadratic loss function defined as:
\begin{align*}
    L= \mathbb{E}[ (f(X) -\kappa(X))^2],
\end{align*}
where $X$ is a uniform random variable on $[0,1]^n$.
Using a stochastic gradient algorithm with the ADAM optimizer, a learning rate of $10^{-3}$, and a batch size of $1000$, we minimize the loss $L$.
The MLPs use a ReLU activation function, with either 2 layers with 10, 20, 40 neurons for each layer or 3 layers with 10, 20, 40, 80, 160 neurons:
\rd{On each plot}, the MLPs are optimized by varying the number of neurons and layers, and only the result that gives the smallest loss during the iterations is kept for the plots.
The different KAN networks are compared using the same parametrization (number of hidden layers, number of neurons, number of meshes used for the 1D functions). The ReLU-KAN has an additional parameter $k$, which we keep at 3 as suggested in the original article.  
\rd{We do not attempt to optimize the number of neurons used in the different KANs: this value is fixed to 10 and is kept relatively low in order to limit the number of parameters. Regarding the mesh size used by the KANs, depending on the network, the default value in the original implementation or in the corresponding articles typically ranges between 5 and 15. Therefore, we choose to test mesh sizes \(P\) belonging to the set \(\{5, 10, 20\}\).}
For all plots, every 100 gradient iterations, the loss is calculated more accurately using $10^5$ samples, giving a series of log-losses plotted using a moving average window of \rd{ 30 }results \rd{depending on the dimension}.\\
ReLU-KAN is very efficient in terms of computation time as it can be broken down into a few operations involving only the ReLU function, matrix addition, and multiplication.
On an 11th generation Intel(R) Core(TM) i7-11850H @ 2.50GHz, using the same parametrization of the KAN nets, the P1-KAN with adaptation computation time is between 1.5 and 2 times slower than the ReLU-KAN, while the P1-KAN without adaptation gives similar computing times.
For the spline version of the KAN originally from \cite{liu2024kan}, we use \href{https://github.com/Blealtan/efficient-KANh}{the efficient Pytorch KAN implementation.}
For the Fast-KAN \cite{li2024kolmogorov}, we use the \href{https://github.com/ZiyaoLi/fast-KAN}{Pytorch implementation.}

\subsection{Results for the A Function}
\begin{figure}[H]
 \begin{minipage}[b]{0.49\linewidth}
  \centering
 \includegraphics[width=\textwidth]{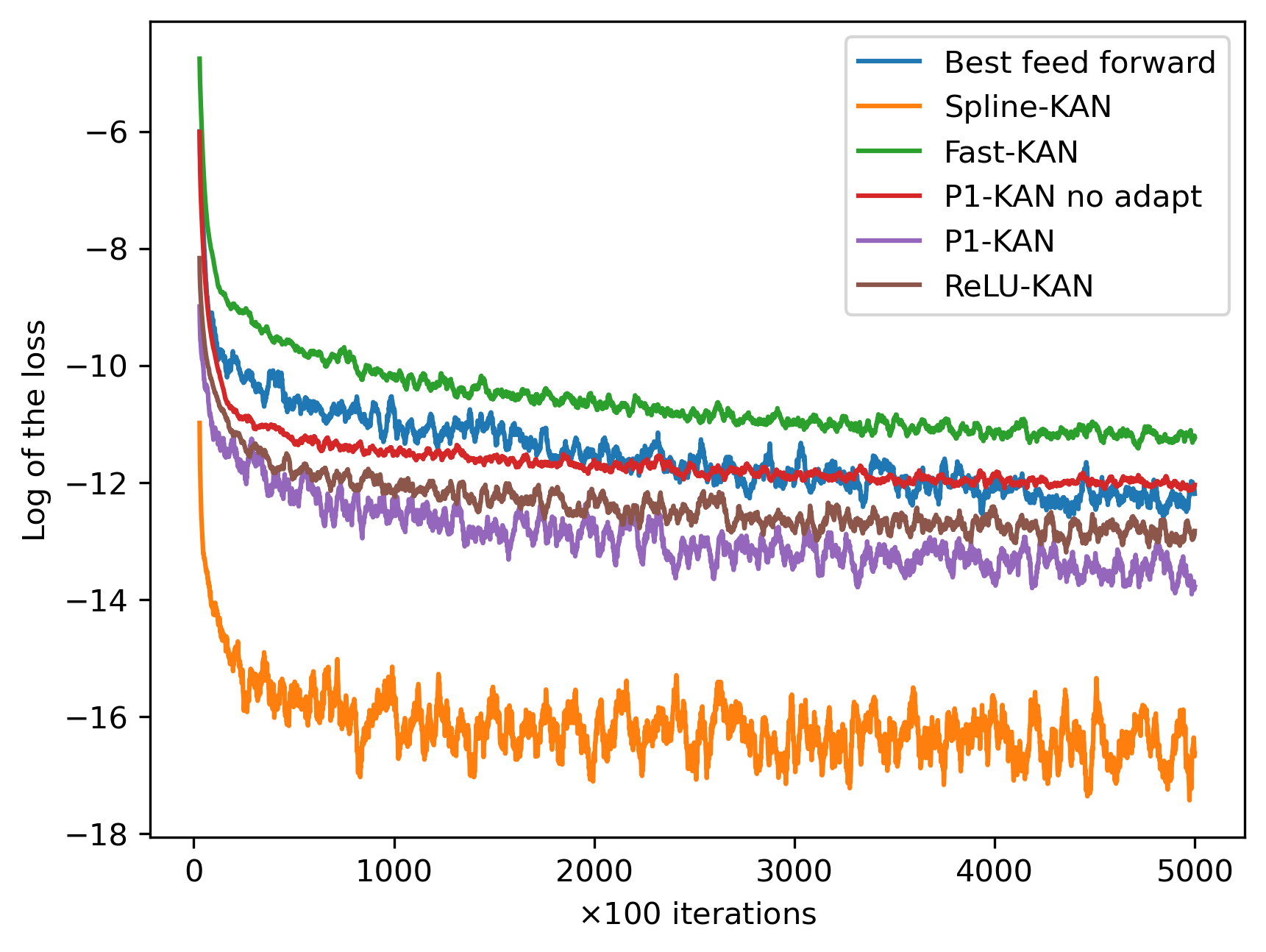}
 \caption*{\tiny 2 hidden layers of 10 neurons, $P=5$.}
 \end{minipage}
 \begin{minipage}[b]{0.49\linewidth}
  \centering
 \includegraphics[width=\textwidth]{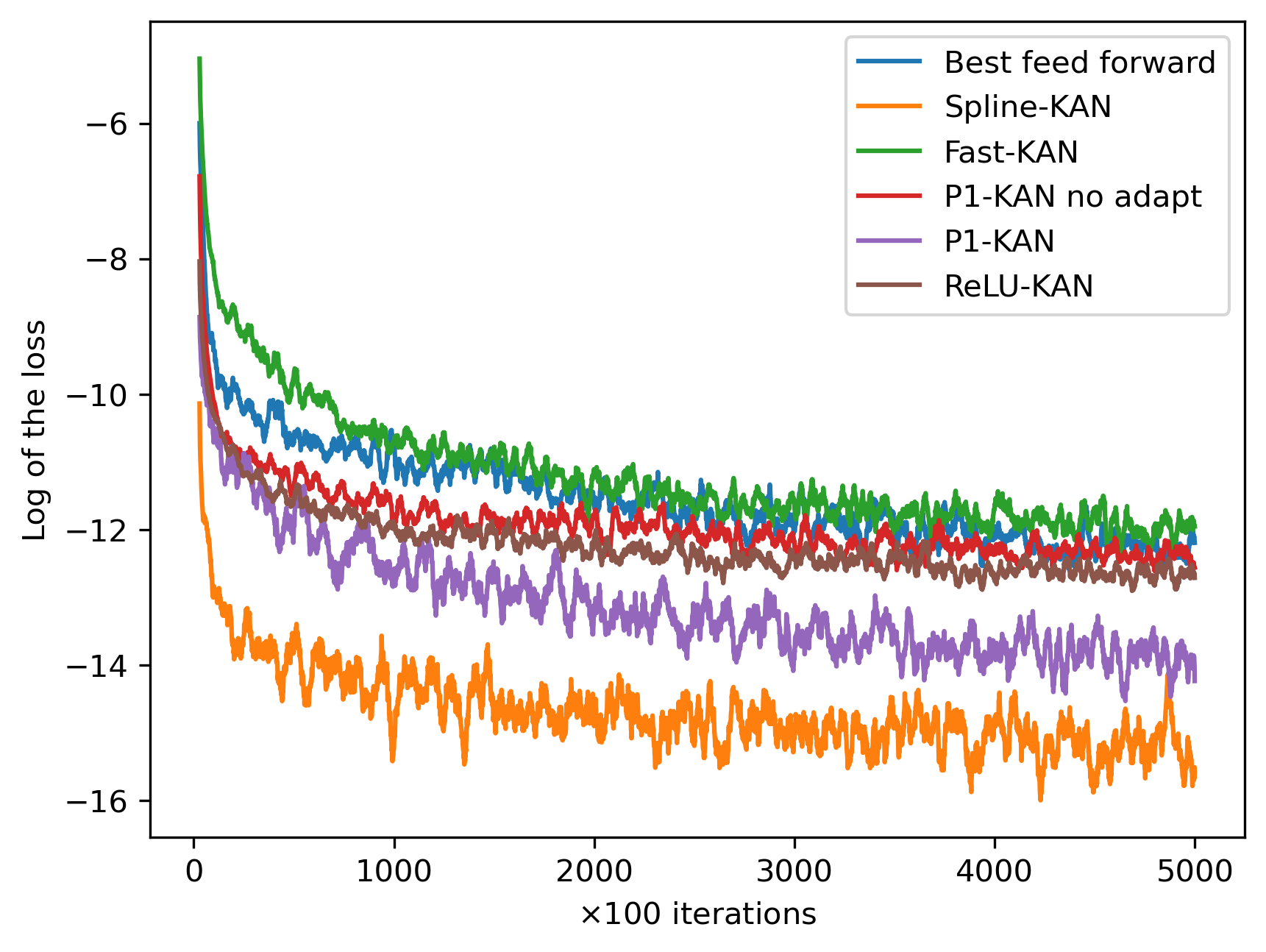}
 \caption*{\tiny 3 hidden layers of 10 neurons, $P=5$.}
 \end{minipage}
 \begin{minipage}[b]{0.49\linewidth}
  \centering
 \includegraphics[width=\textwidth]{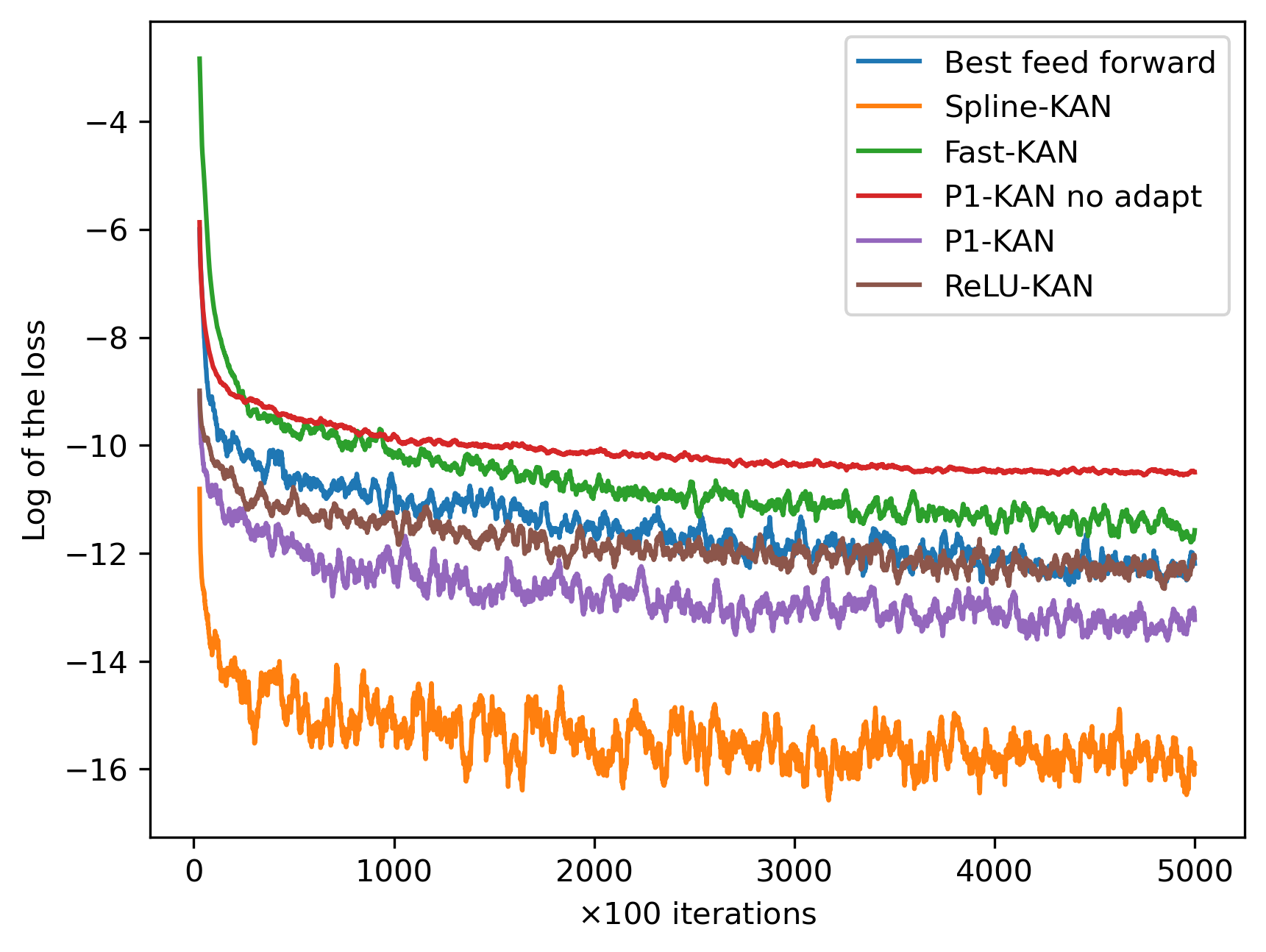}
 \caption*{\tiny 2 hidden layers of 10 neurons, $P=10$.}
 \end{minipage}
  \begin{minipage}[b]{0.49\linewidth}
  \centering
 \includegraphics[width=\textwidth]{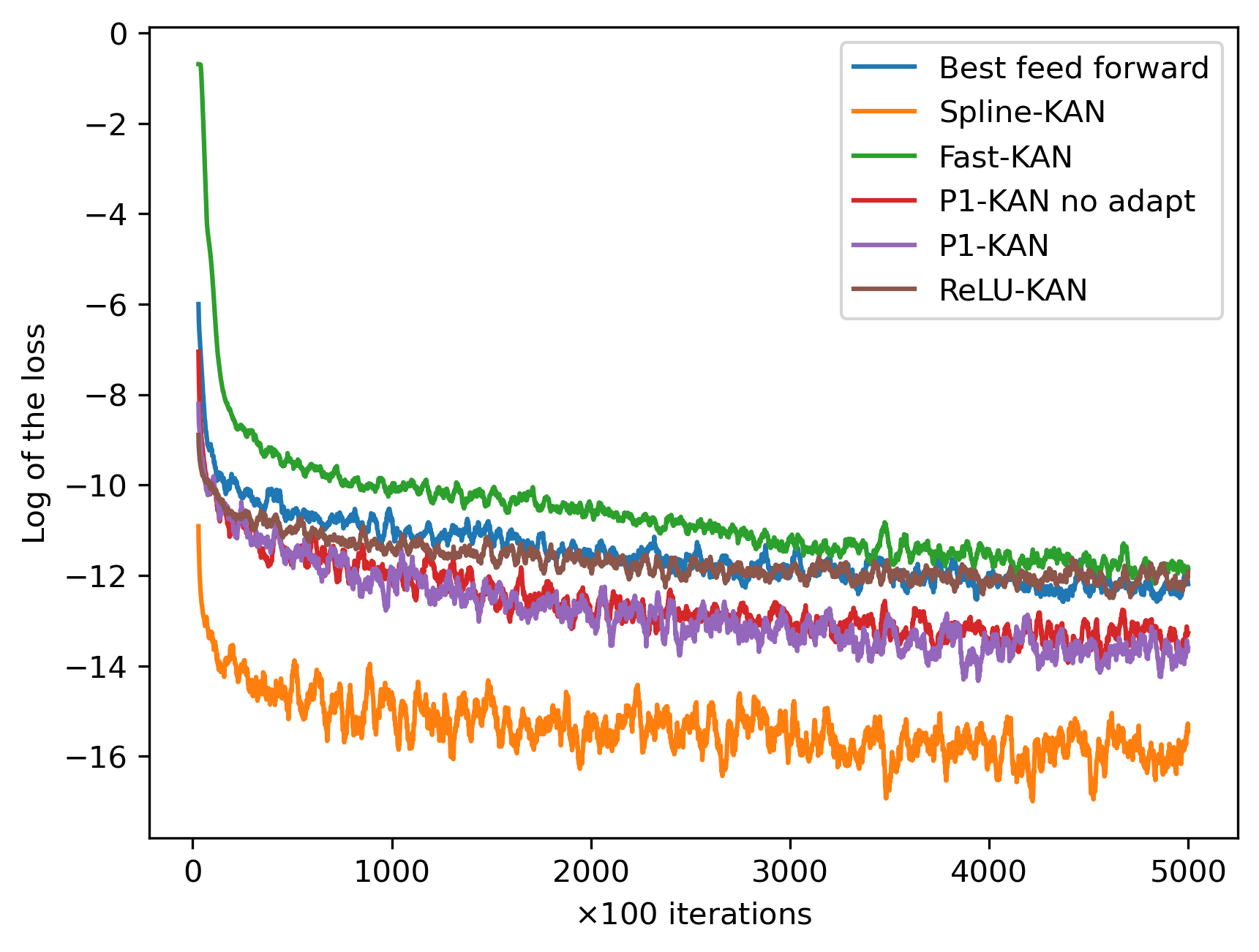}
 \caption*{\tiny 3 hidden layers of 10 neurons, $P=10$.}
 \end{minipage}
  \begin{minipage}[b]{0.49\linewidth}
  \centering
 \includegraphics[width=\textwidth]{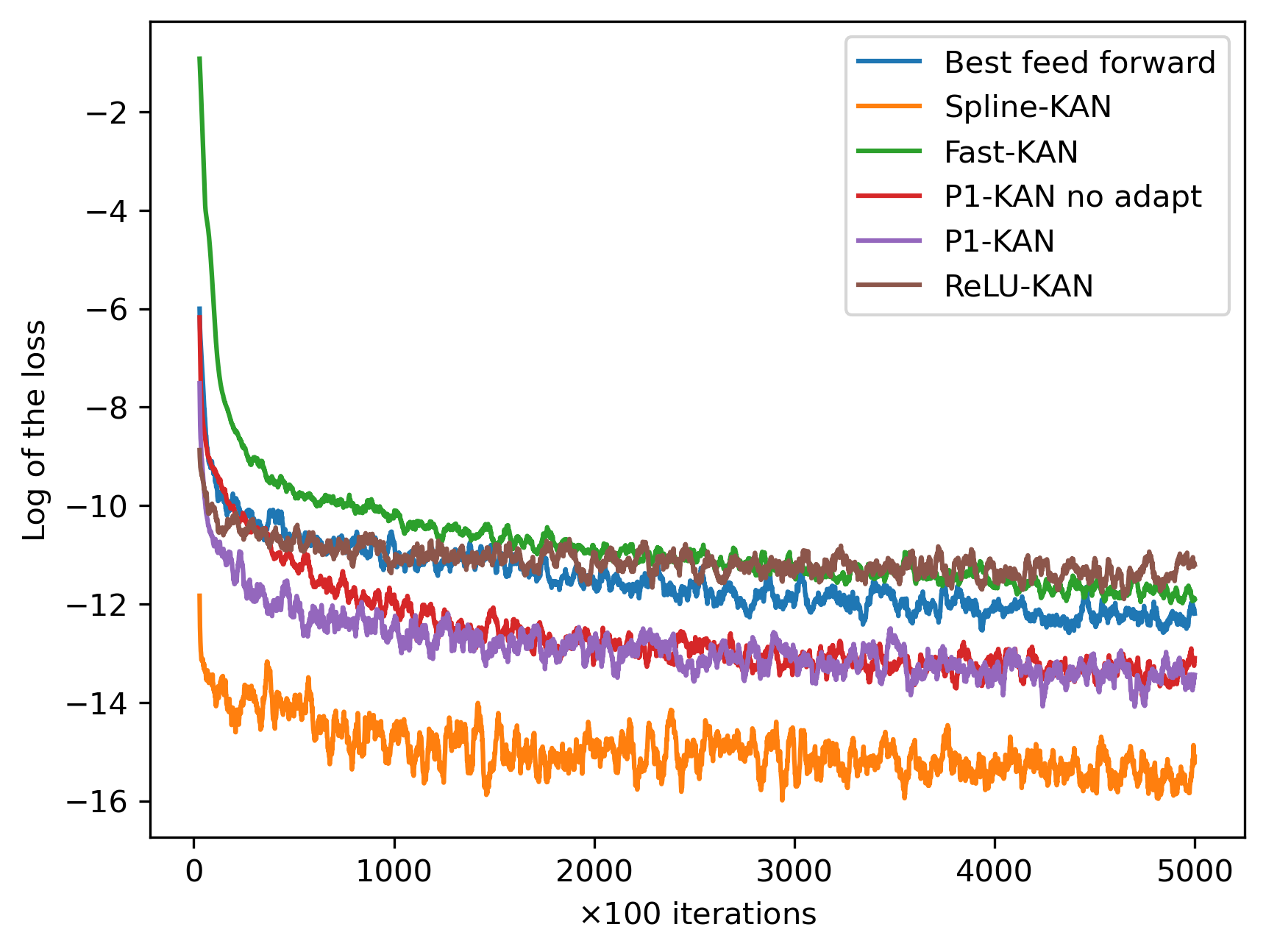}
 \caption*{\tiny 2 hidden layers of 10 neurons, $P=20$.}
 \end{minipage}
 \begin{minipage}[b]{0.49\linewidth}
  \centering
 \includegraphics[width=\textwidth]{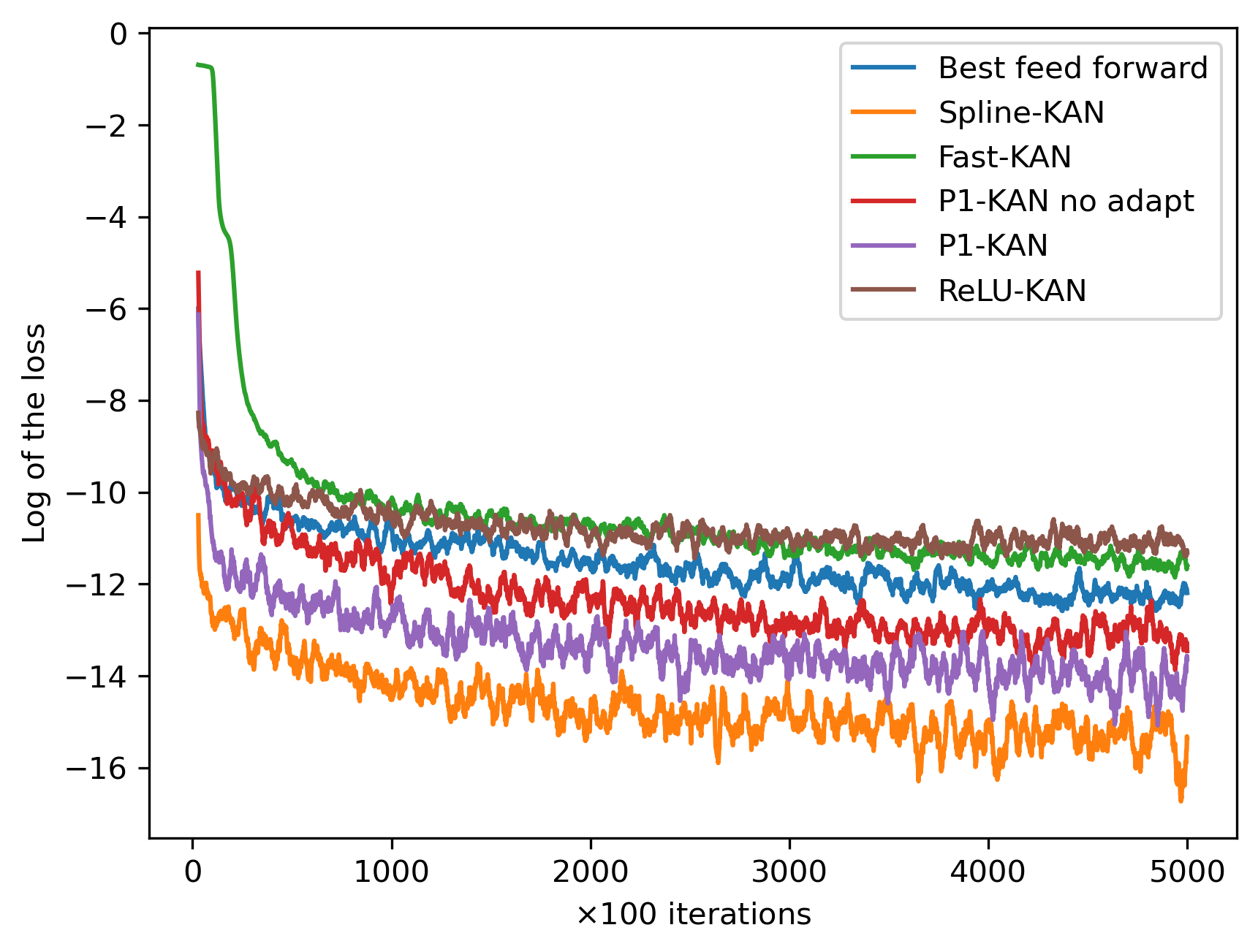}
 \caption*{\tiny 3 hidden layers of 10 neurons, $P=20$.}
 \end{minipage}
  \caption{Results in dimension 6 for function A \rd{for a given number of hidden layers and a given $P$, $10$ neurons taken : superiority of the Spline-KAN for smooth functions is observed}. \label{fig:A6}}
\end{figure}

\begin{figure}[H]
 \begin{minipage}[b]{0.49\linewidth}
  \centering
 \includegraphics[width=\textwidth]{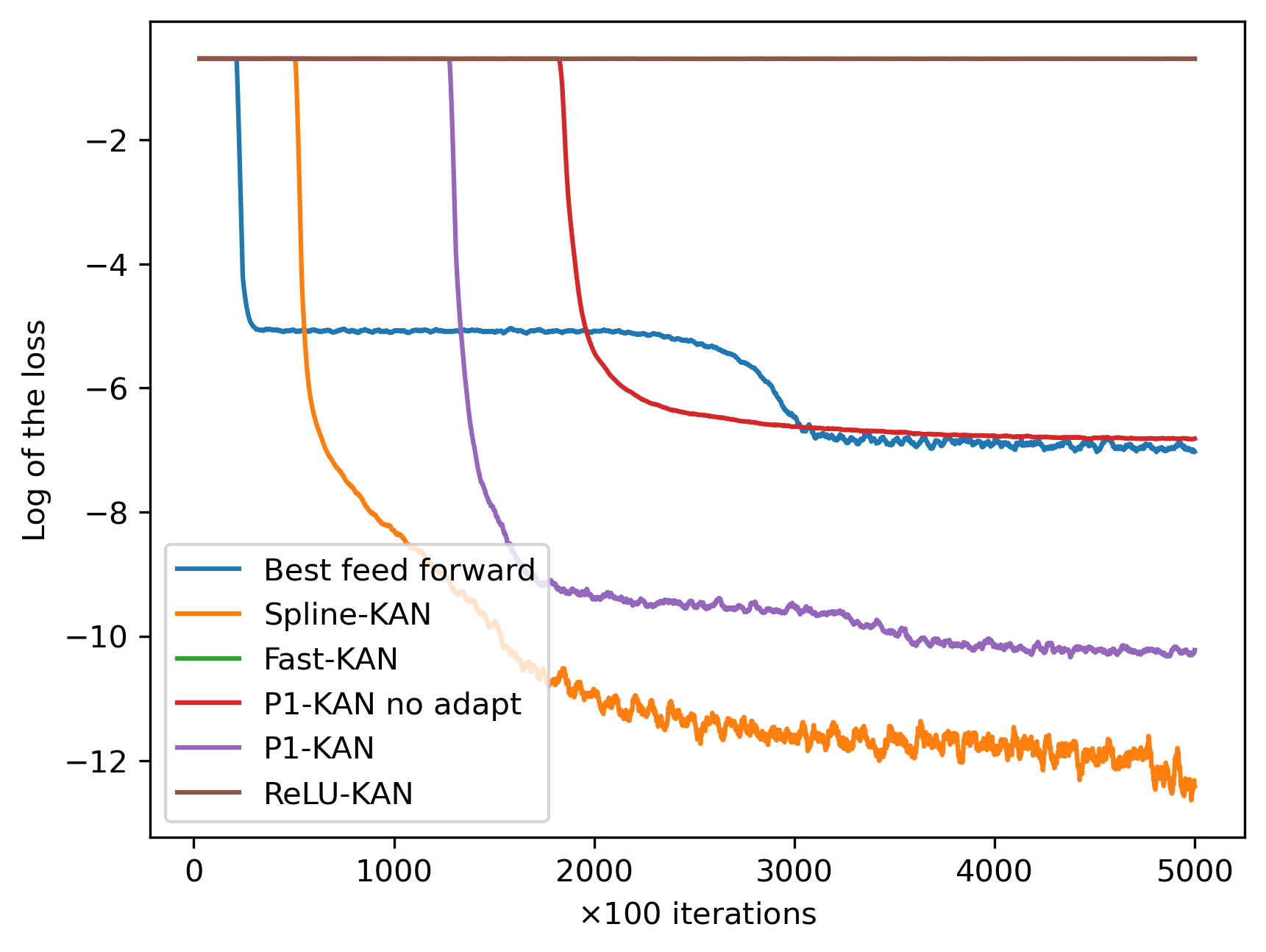}
 \caption*{\tiny 2 hidden layers of 10 neurons, $P=5$.}
 \end{minipage}
 \begin{minipage}[b]{0.49\linewidth}
  \centering
 \includegraphics[width=\textwidth]{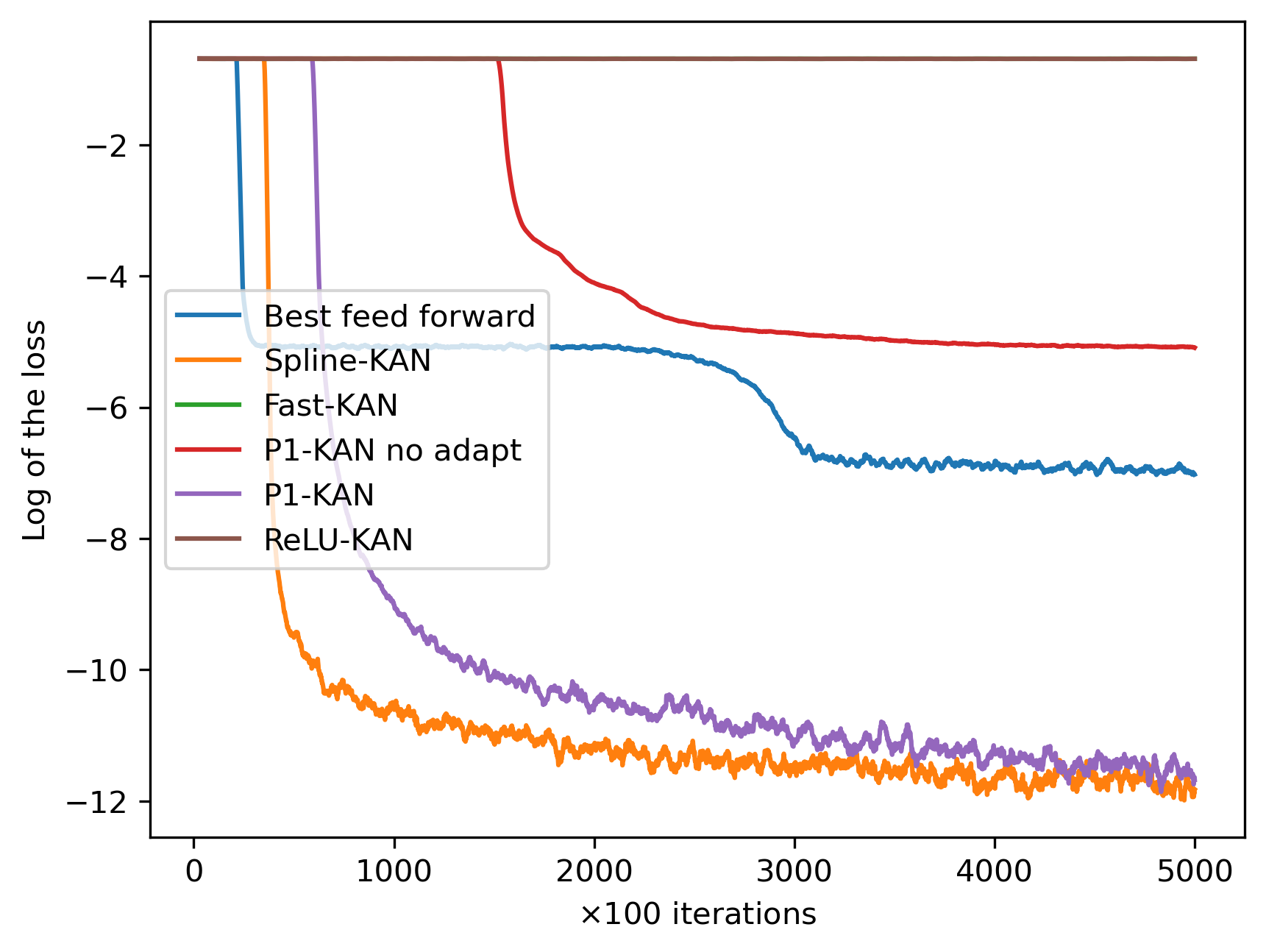}
 \caption*{\tiny 3 hidden layers of 10 neurons, $P=5$.}
 \end{minipage}
  \begin{minipage}[b]{0.49\linewidth}
  \centering
 \includegraphics[width=\textwidth]{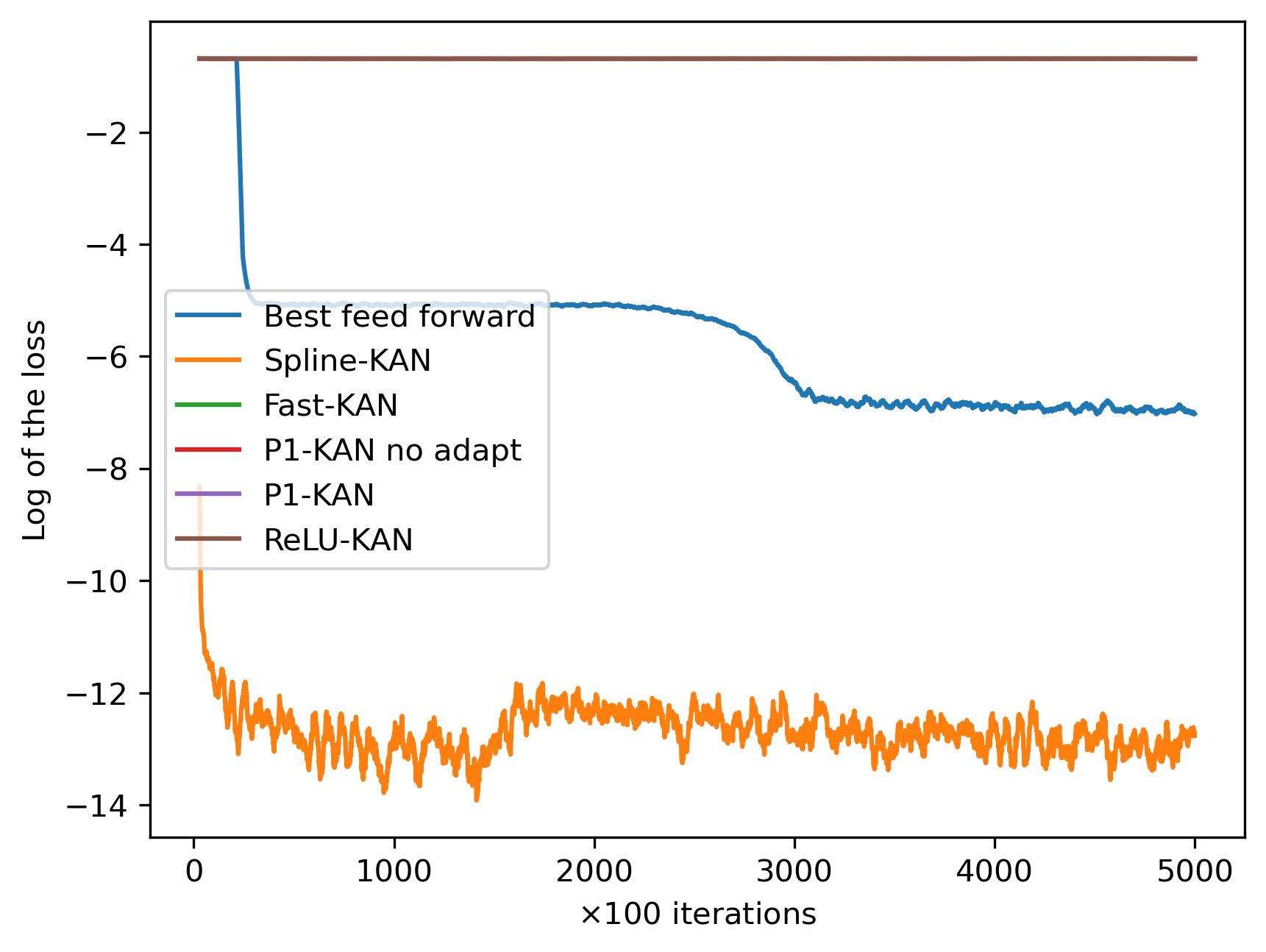}
 \caption*{\tiny 2 hidden layers of 10 neurons, $P=10$.}
 \end{minipage}
 \begin{minipage}[b]{0.49\linewidth}
  \centering
 \includegraphics[width=\textwidth]{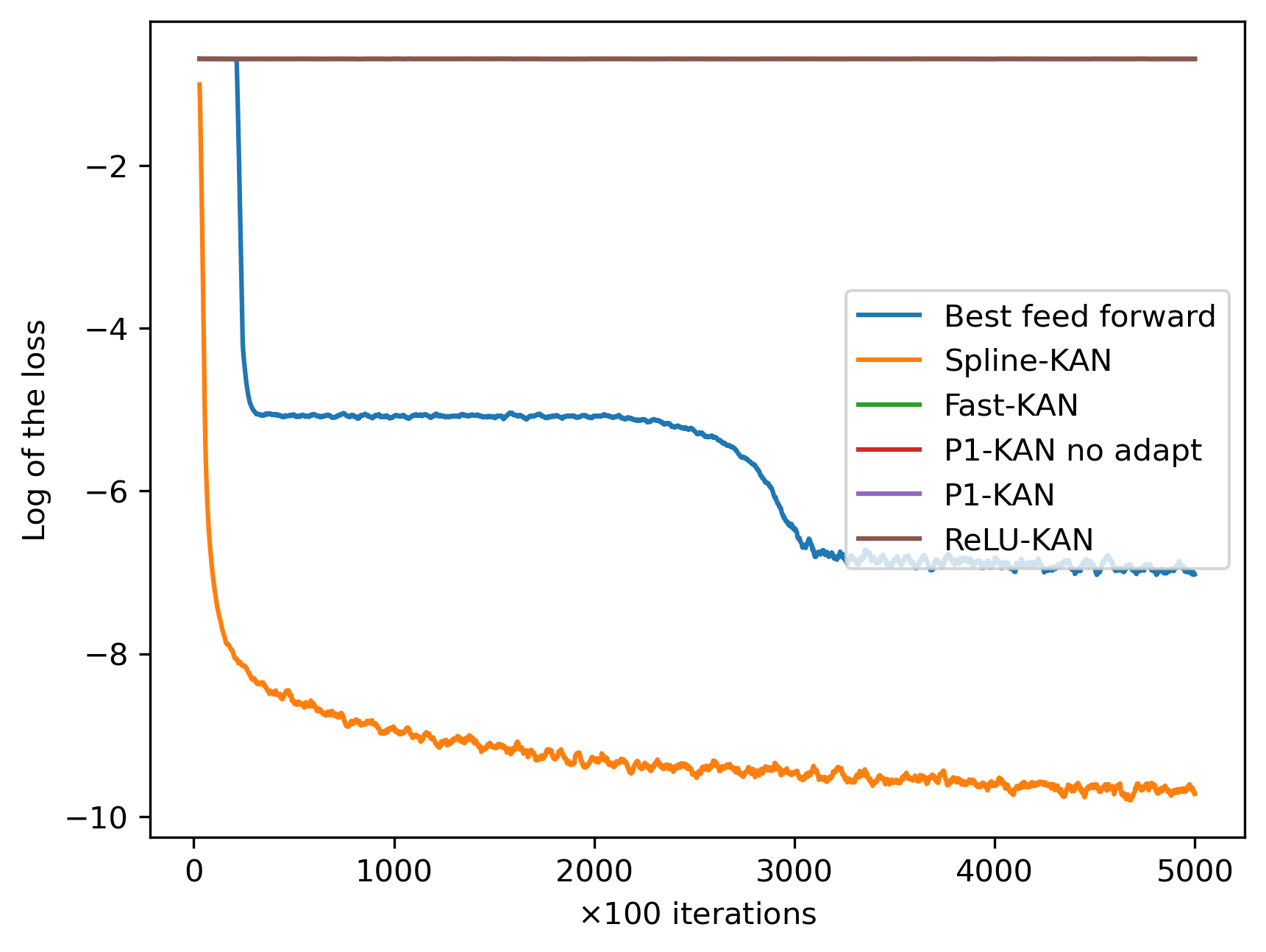}
 \caption*{\tiny 3 hidden layers of 10 neurons, $P=10$.}
 \end{minipage}
  \caption{Results in dimension 12 for function A \rd{for a given number of hidden layers and a given $P$, $10$ neurons taken : with $P$ low, P1-KAN and Spline-KAN have same accuracy .} \label{fig:A12}}
\end{figure}

The results in dimension 6 shown in Figure \ref{fig:A6} on a single run seem to indicate that the original Spline-KAN network converges faster than the P1-KAN network with adaptation, which is the second most effective network. 
In general, the ReLU-KAN network converges at least as well as the best feedforward network, while the Fast-KAN is the least effective network.
The P1-KAN without adaptation is less effective than the P1-KAN with adaptation.
In dimension 12, as shown in Figure \ref{fig:A12}, the Spline-KAN and the P1-KAN with adaptation again give the best results. The Fast-KAN and the ReLU-KAN fail, while the feedforward network seems to converge very slowly and its accuracy remains limited.  Again, adaptation of the P1-KAN gives far better results.

\rd{
As these graphs are obtained from a single run, Table~\ref{tab:compNetRegular} reports 
the average results and standard deviations computed over 10 runs of the different 
networks in dimensions 11 and 12, each trained for $10^6$ stochastic gradient iterations.
For the MLP, we test two configurations with either 2 or 3 hidden layers. 
When a network fails to converge, the best approximation it can produce is the 
constant function, and the MSE remains equal to $0.5$.
Therefore, an average value between $0.1$ and $0.5$ indicates that at least one 
run did not converge. We highlight in bold the configurations that successfully 
converged in all 10 runs.
Dimension 12 represents a threshold: beyond this dimension, all methods fail.
\begin{itemize}
    \item In dimension 12, some of the 10 MLP runs did not converge. Most Fast-KAN and ReLU-KAN configurations also 
    failed to converge. The Spline-KAN converged in all configurations using $5$ or 
    $10$ meshes and appears to give the best results overall.
    The P1-KAN, with and without adaptation, converged in all 10 runs when the number 
    of meshes was small.
    \item In dimension 11, most of the networks managed to converge for all the 10 runs. ReLU-KAN achieves an accurate result with $P=5$ but fails with $P=10$. Overall, Spline-KAN remains the best network followed by the P1-KAN.
\end{itemize}
Clearly, adaptation leads to better performance for the P1-KAN network.
}

\begin{table}[H]
    \centering
    \begin{tabular}{|c|c|c|c|c|c|c|c|} \hline
     &          &             &      &  \multicolumn{2}{c|}{ Dim = 11} & \multicolumn{2}{c|}{Dim = 12} \\ \cline{5-8}
     Method & Nb  Layers &  Nb Neurons & $P$  & Average & Std  & Average & Std \\ \hline
MLP   &  2  &  160  &  &       {\bf 5.35E-05}  &  4.80E-05  &  1.37E-01  &  2.11E-01  \\
MLP   &  3  &  160  &   &    {\bf  3.10E-05}  &  1.95E-05 &  1.50E-01  & 2.29E-01  \\ \hline
Spline-KAN   &  2  &  10  &    5  &  {\bf 1.92E-05}  &  3.29E-05   &  {\bf 1.91E-04}  &  4.56E-04\\
Spline-KAN   &  2  &  10  &     10  & {\bf  2.78E-05}  &  4.04E-05 &  {\bf 1.52E-05}  &  2.77E-05 \\ \hline
ReLU-KAN   &  2  &  10  & 5    &       {\bf 4.69E-05}  &  3.21E-05 &  1.00E-01  &  2.00E-01\\
ReLU-KAN   &  2  &  10  & 10  &  4.50E-01  &  1.49E-01  &   5.00E-01  &  1.65E-03 \\ \hline
Fast-KAN   &  2  &  10  & 5  &   {\bf 3.78E-04}  &  3.00E-04 &  2.50E-01  &  2.49E-01\\
Fast-KAN   &  2  &  10  & 10  &  {\bf 3.01E-03}  &  5.68E-03  &  4.00E-01  &  1.99E-01 \\ \hline
P1-KAN   &  2  &  10  & 5  &             {\bf 1.71E-04}  &  2.70E-04                &  {\bf 9.01E-05}  &  8.16E-05 \\
P1-KAN   &  2  &  10  & 10  &     {\bf 1.15E-04 }  &  1.13E-04                        &2.50E-01  &  2.50E-01 \\ \hline
P1-KAN no adapt   &  2  &  10  & 5  &     {\bf 4.69E-03}  &  8.68E-03     &  {\bf 8.87E-03}  &  1.45E-02 \\
P1-KAN no adapt   &  2  &  10  & 10  &  {\bf 1.91E-03}  &  2.15E-03 &   3.52E-01  &  2.27E-01  \\ \hline

\end{tabular} 
    \caption{\rd{Averaged  and standard deviation of  MSE obtained from 10 runs  in dimension 11 and 12 for function A: Spline-KAN followed by  P1-KAN  dominates the test.}}
    \label{tab:compNetRegular}
\end{table}

\rd{
Besides, our Theorem~\ref{theo1} states that with adaptation, even with $P$ as low as 2, we can approximate our function as closely as desired.
In Table~\ref{tab:dimLowP}, we report the average MSE and the standard deviation obtained for $P$ varying from 2 to 5. 
These results support the theoretical findings: even taking $P = 2$ yields good performance, and the accuracy generally increases with larger values of $P$.
\begin{table}[H]
    \centering
    \begin{minipage}{0.45\textwidth}
    \centering
    \begin{tabular}{|c|c|c|} \hline
        $P$ & Average & std \\ \hline
        2  &  5.47E-06  &  4.72E-06 \\
        3 &   2.19E-06  &  1.78E-06   \\
        4 &   1.14E-06  &  6.22E-07  \\
        5 &   1.42E-06  &  1.09E-06   \\ \hline
    \end{tabular}
    \caption*{Dimension 3}
      \end{minipage}
          \begin{minipage}{0.45\textwidth}
    \centering
       \begin{tabular}{|c|c|c|} \hline
        $P$ & Average & std \\ \hline
        2  &   6.71E-05  &  4.96E-05 \\
        3 &    7.86E-06  &  3.51E-06 \\
        4 &    6.15E-06  &  5.91E-06 \\
        5 &     3.09E-06  &  1.72E-06 \\ \hline
    \end{tabular}
    \caption*{Dimension 6}
      \end{minipage}
      \caption{\rd{Averaged MSE and standard deviation associated for low values of $P$  when adaptation is used in  P1-KAN  with 2 hidden layers of 10 neurons (function A): taking $P=2$ converges but a lower accuracy is observed. } \label{tab:dimLowP}}
\end{table}
}

\subsection{Results for the B Function}
\begin{figure}[H]
 \begin{minipage}[b]{0.49\linewidth}
  \centering
 \includegraphics[width=\textwidth]{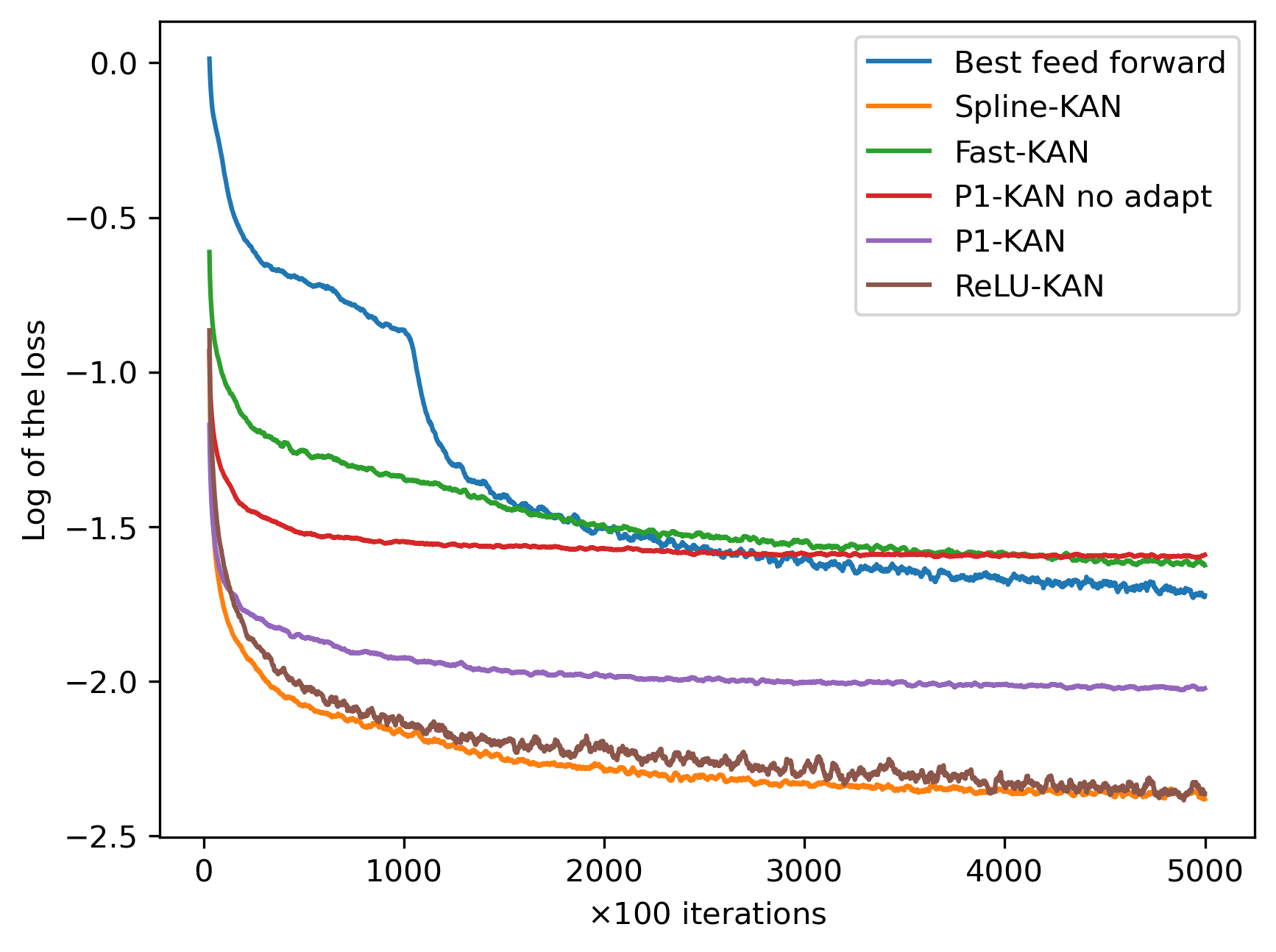}
 \caption*{\tiny 2 hidden layers of 10 neurons, $P=5$.}
 \end{minipage}
 \begin{minipage}[b]{0.49\linewidth}
  \centering
 \includegraphics[width=\textwidth]{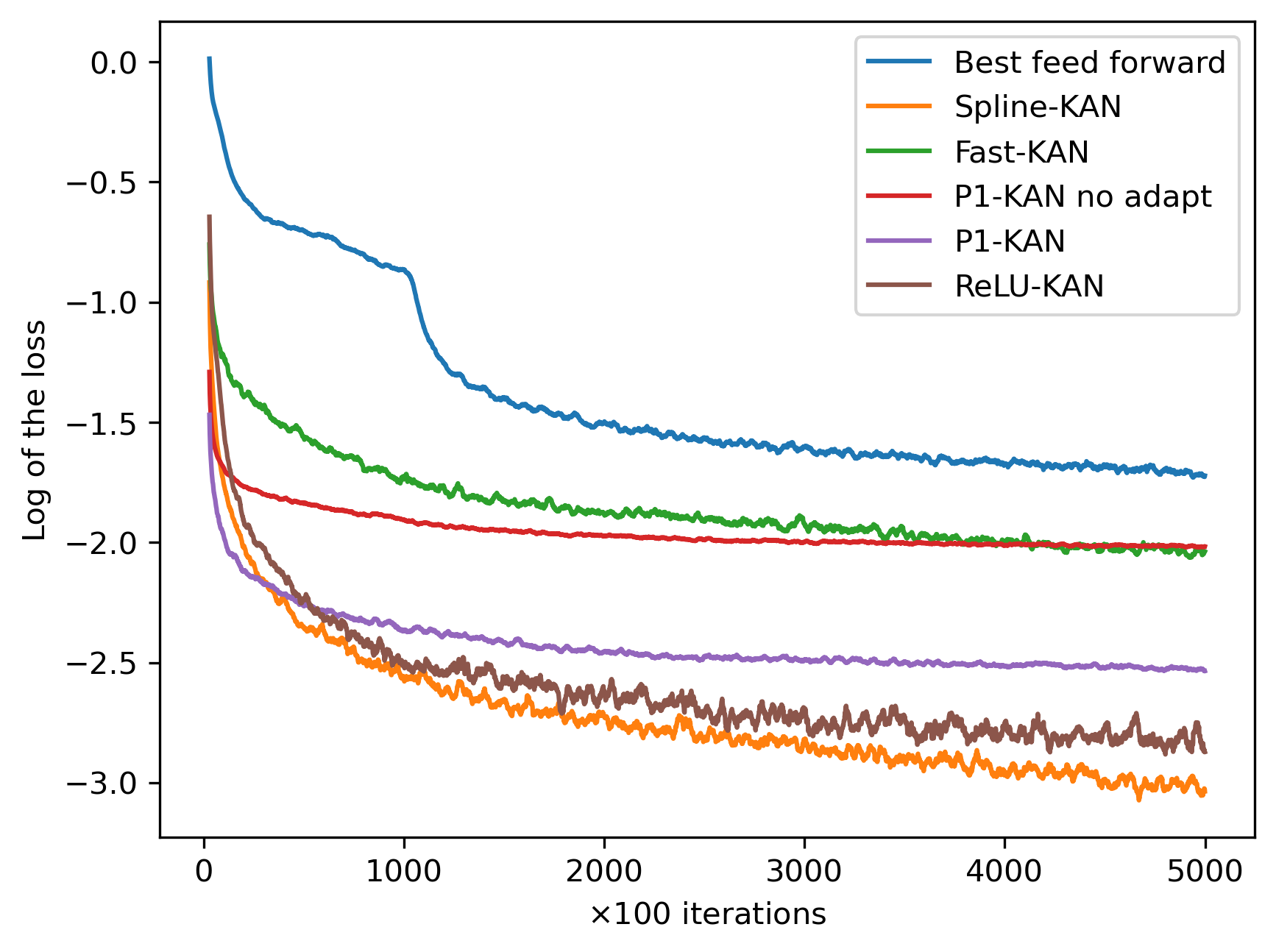}
 \caption*{\tiny 3 hidden layers of 10 neurons, $P=5$.}
 \end{minipage}
 \begin{minipage}[b]{0.49\linewidth}
  \centering
 \includegraphics[width=\textwidth]{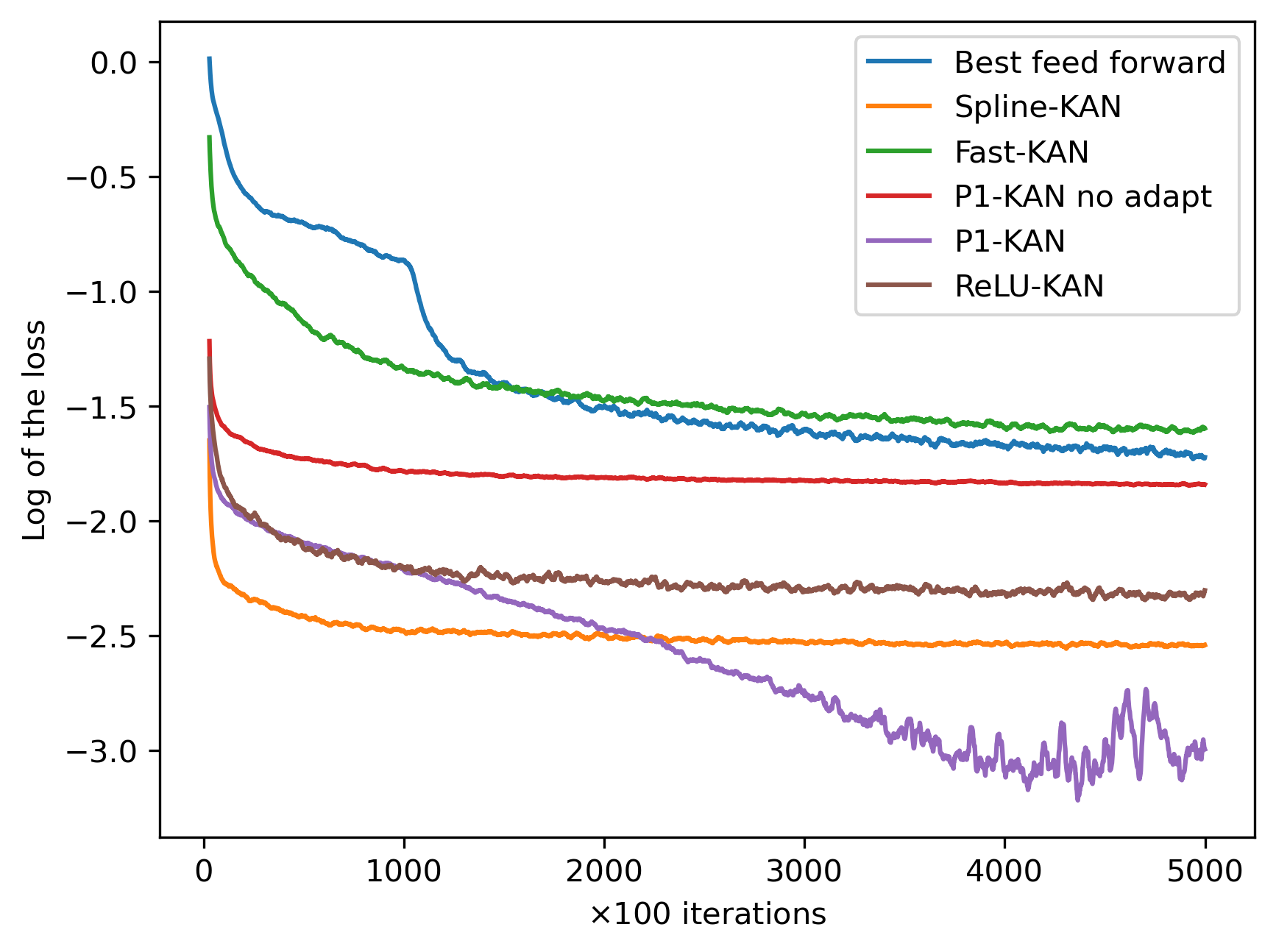}
 \caption*{\tiny 2 hidden layers of 10 neurons, $P=10$.}
 \end{minipage}
  \begin{minipage}[b]{0.49\linewidth}
  \centering
 \includegraphics[width=\textwidth]{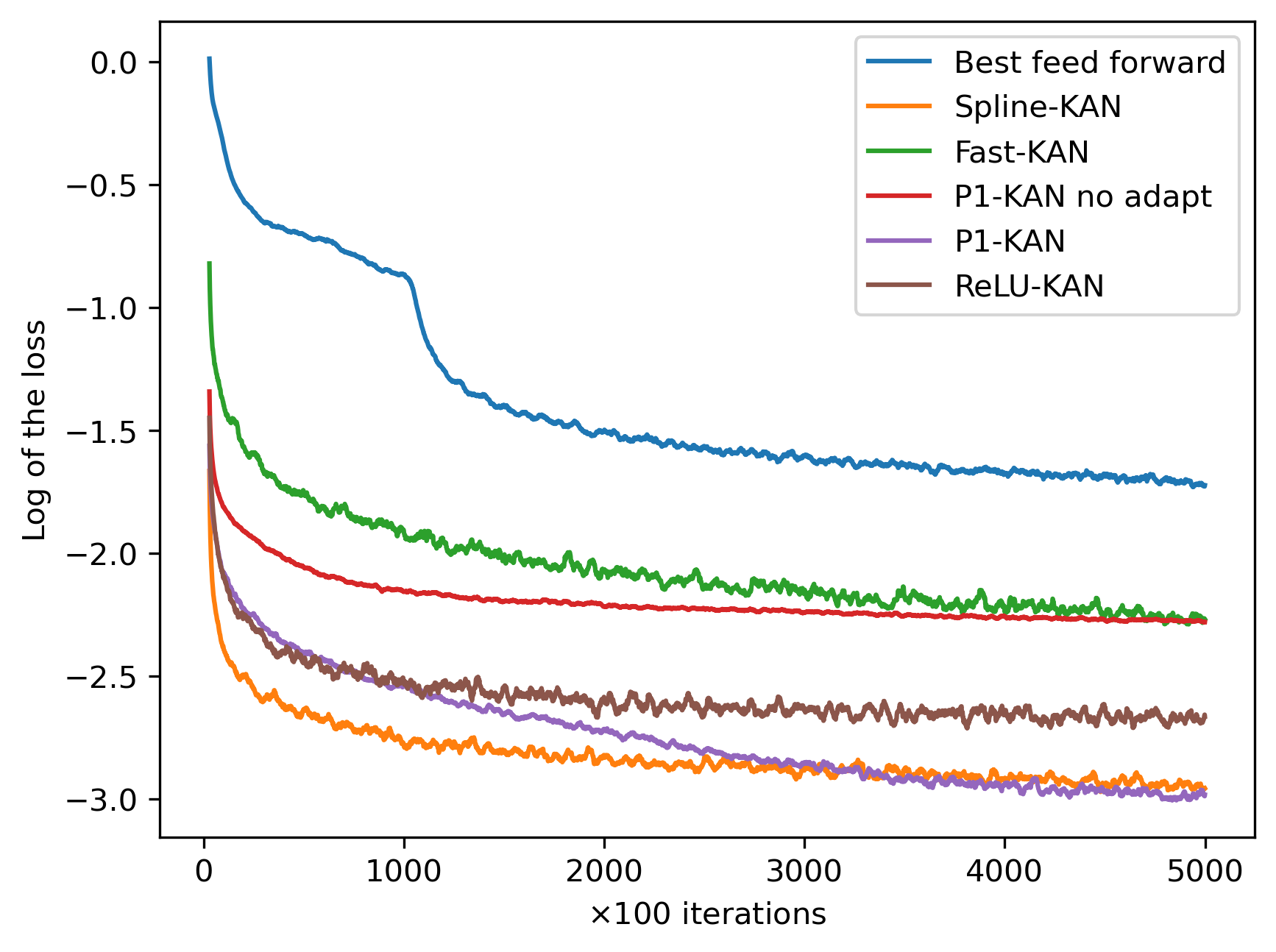}
 \caption*{\tiny 3 hidden layers of 10 neurons, $P=10$.}
 \end{minipage}
  \begin{minipage}[b]{0.49\linewidth}
  \centering
 \includegraphics[width=\textwidth]{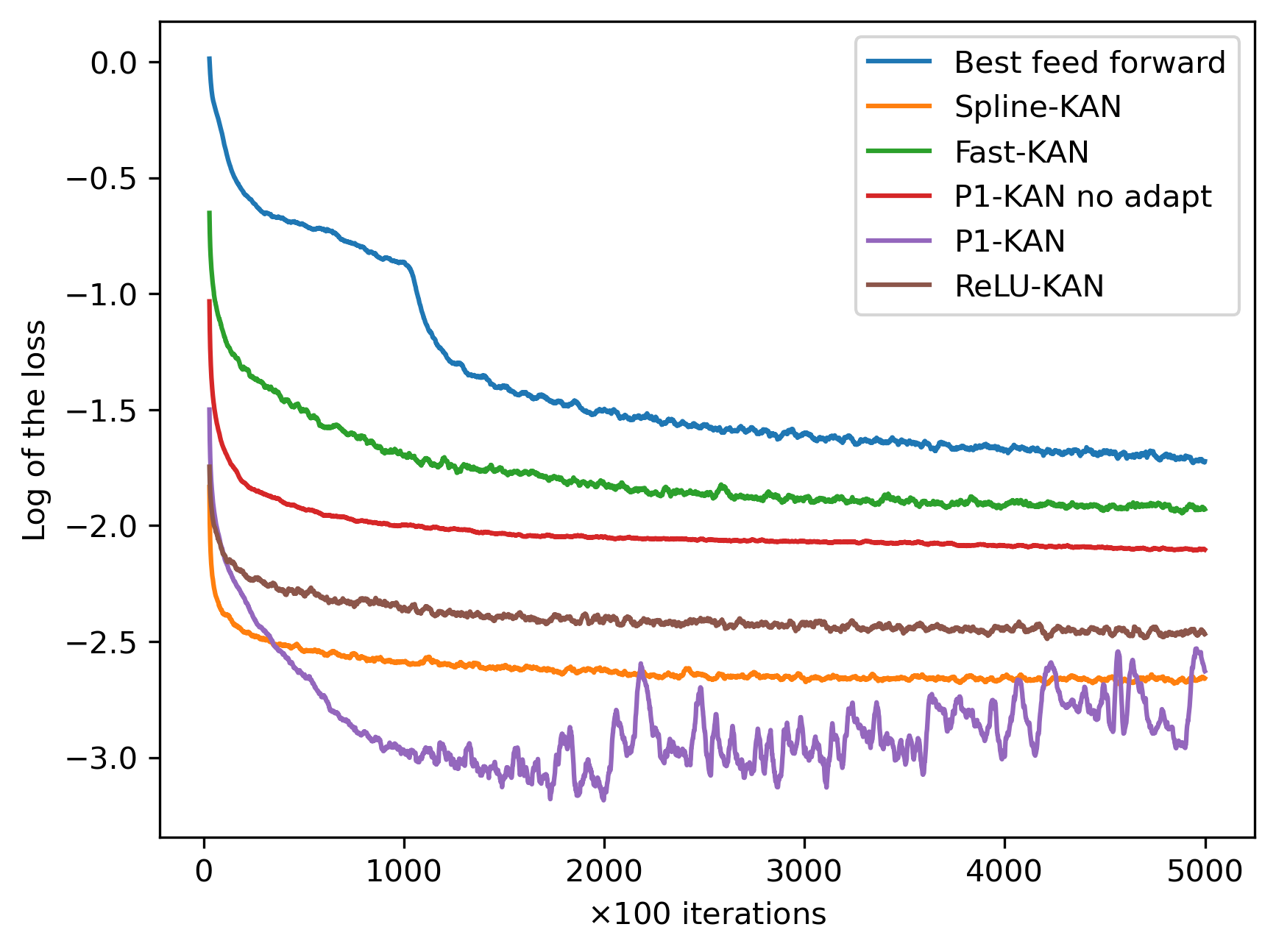}
 \caption*{\tiny 2 hidden layers of 10 neurons, $P=20$.}
 \end{minipage}
  \begin{minipage}[b]{0.49\linewidth}
  \centering
 \includegraphics[width=\textwidth]{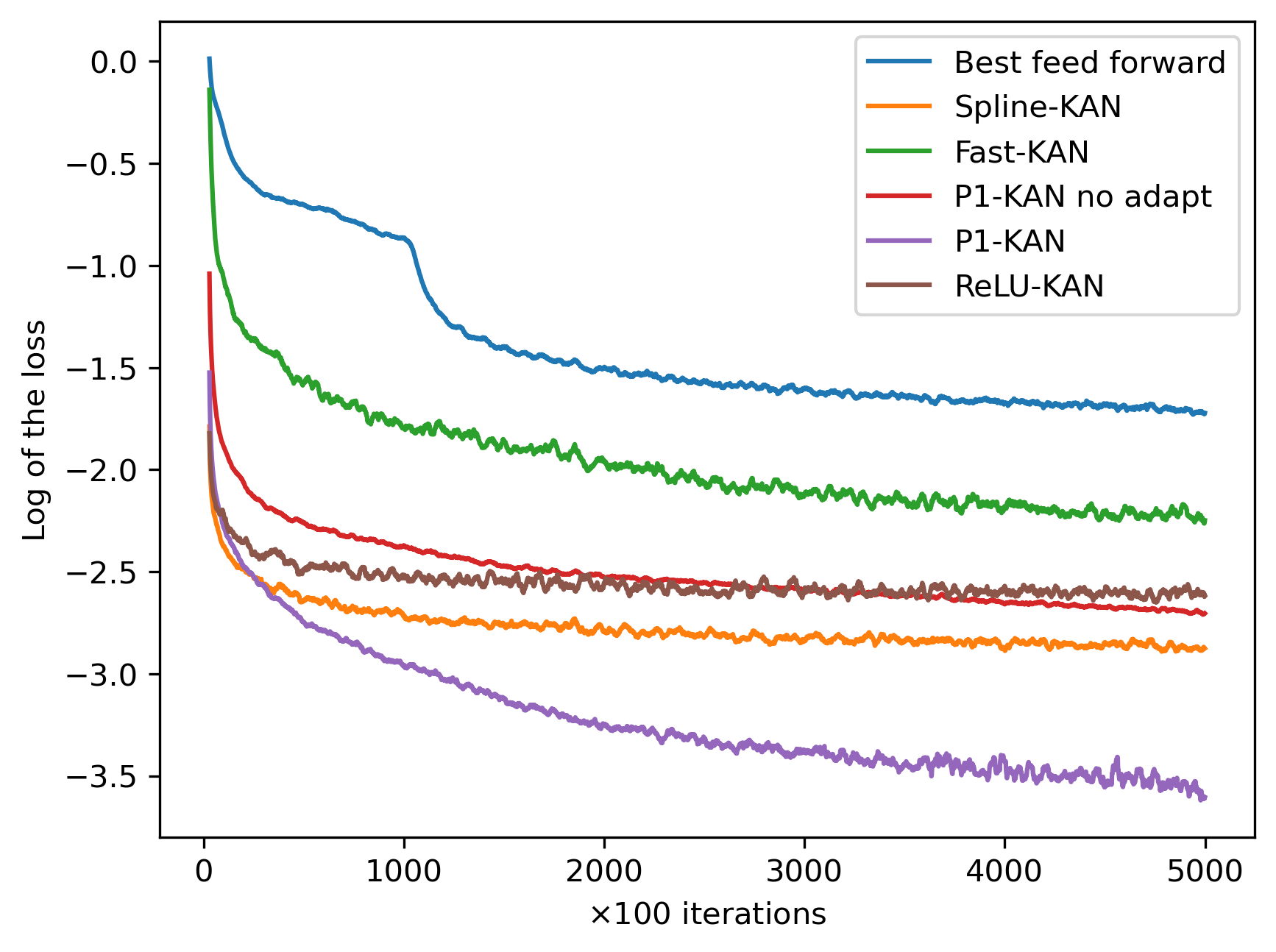}
 \caption*{\tiny 3 hidden layers of 10 neurons, $P=20$.}
 \end{minipage}
  \caption{Results in dimension 2 for function B \rd{for a given number of hidden layers and a given $P$, $10$ neurons taken : irregular functions favor P1-KAN}. \label{fig:B2}}
\end{figure}

\begin{figure}[H]
 \begin{minipage}[b]{0.49\linewidth}
  \centering
 \includegraphics[width=\textwidth]{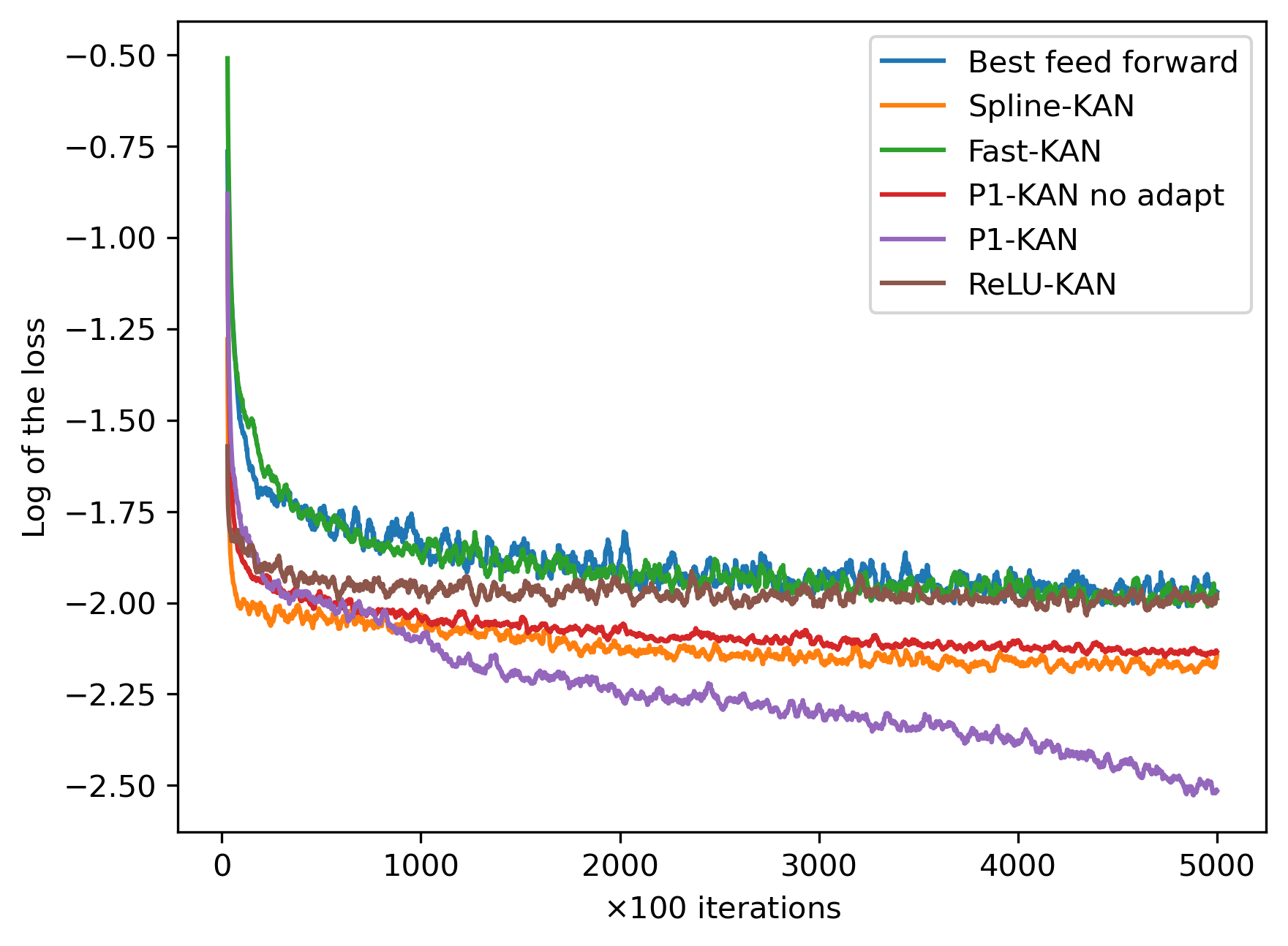}
 \caption*{\tiny 2 hidden layers of 10 neurons, $P=10$}
 \end{minipage}
  \begin{minipage}[b]{0.49\linewidth}
  \centering
 \includegraphics[width=\textwidth]{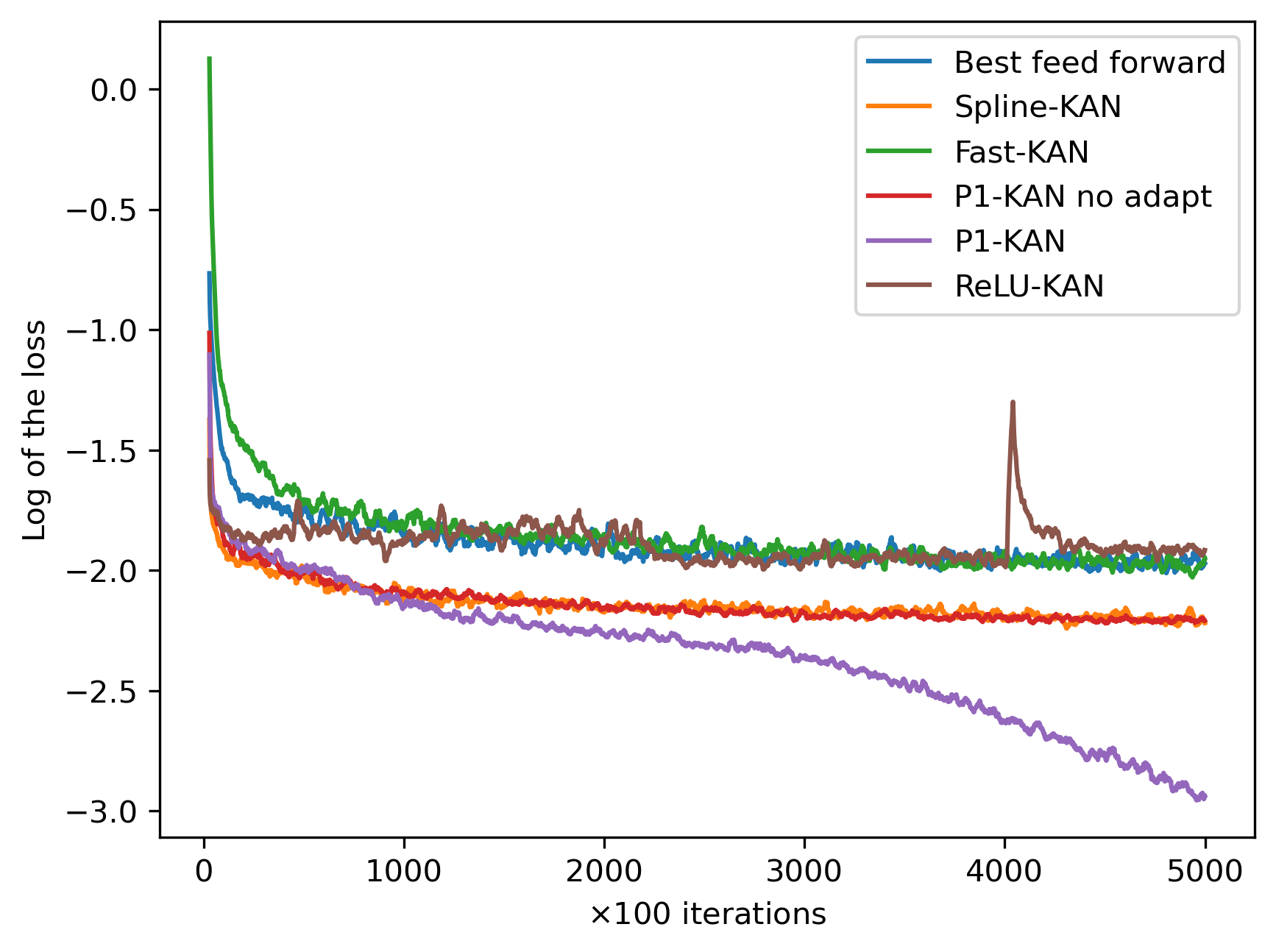}
 \caption*{\tiny 3 hidden layers of 10 neurons, $P=10$}
 \end{minipage}
  \begin{minipage}[b]{0.49\linewidth}
  \centering
 \includegraphics[width=\textwidth]{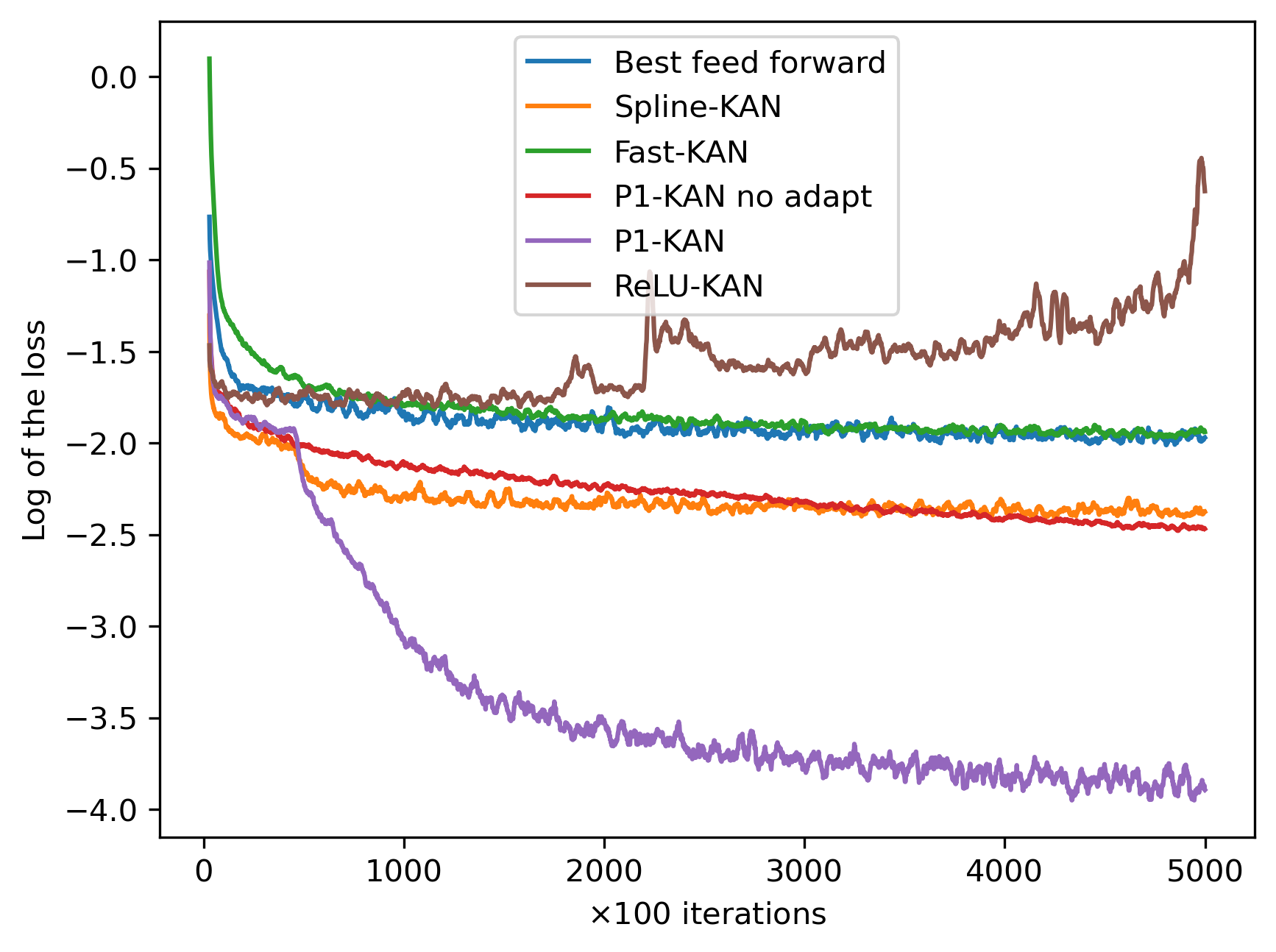}
 \caption*{\tiny 2 hidden layers of 10 neurons, $P=20$}
 \end{minipage}
  \begin{minipage}[b]{0.49\linewidth}
  \centering
 \includegraphics[width=\textwidth]{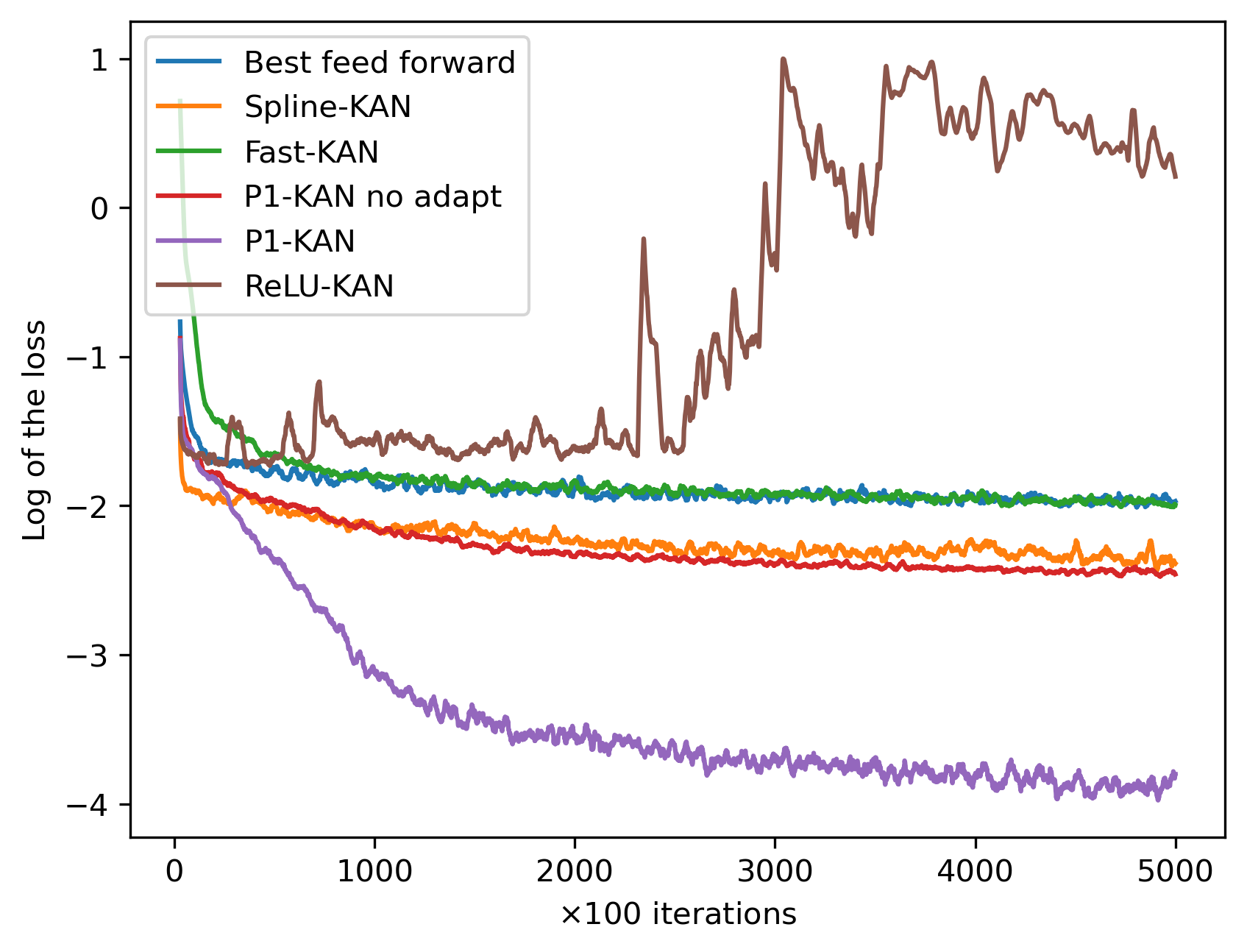}
 \caption*{\tiny 3 hidden layers of 10 neurons, $P=20$}
 \end{minipage}
  \caption{Results in dimension 5 for function B \rd{for a given number of hidden layers and a given $P$, $10$ neurons taken : only P1-KAN optimizes correctly  as the dimension grows.} \label{fig:B5}}
\end{figure}

In dimension 2, as shown in Figure \ref{fig:B2}, we see that the feedforward network lags behind the KAN networks.
By taking high values of $P$, the P1-KAN network is the only network that gives very good results. The Spline-KAN is the second-best network.
Since the B function is irregular, we expected that the best approximation using KAN networks involves the use of steeply varying $\psi$ and $\Phi$ functions. Therefore, we expect the Spline-KAN with a B-spline of order three to be less effective than the linear piecewise approximation of the P1-KAN.
Finally, if we go up to dimension 5 on figure \ref{fig:B5} , we see that the ReLU-KAN network can diverge. The P1-KAN network is the only one that gives acceptable results by using 2 or 3 hidden layers of 10 neurons and $P=20$.

\rd{  The average MSE and the associated standard deviations obtained for function B in dimensions 4 and 5 is given in Table \ref{tab:compP1SplineIrrg}.
Results shown in bold indicate an average MSE below $0.1$; the P1-KAN achieves the best performance.
Once again, we observe that ReLU-KAN diverges when $P$ is too large.
As in the regular case, adapting the P1-KAN improves accuracy.}
\begin{table}[H]
    \centering
    \begin{tabular}{|c|c|c|c|c|c|c|c|} \hline
     &          &             &      &  \multicolumn{2}{c|}{ Dim = 4} & \multicolumn{2}{c|}{Dim = 5} \\ \cline{5-8}
Method & Nb  Layers &  Nb Neurons & $P$  & Average & Std  & Average & Std \\ \hline   
MLP   &  2  &  160  &  &  2.50E-01  &  1.79E-02  &  1.37E-01  &  7.36E-03\\
MLP   &  3  &  160  &   &  2.00E-01  &  8.12E-03   &  1.25E-01  &  6.47E-03 \\ \hline
Spline-KAN   &  2  &  10  & 5  &  2.00E-01  &  7.88E-03  &  1.10E-01  &  2.37E-03 \\
Spline-KAN   &  2  &  10  & 10 &  1.56E-01  &  1.64E-02 &  1.07E-01  &  1.09E-02\\
Spline-KAN   &  2  &  10  & 20  &  1.11E-01  &  1.75E-02  &  {\bf 8.45E-02}  &  1.30E-02 \\  \hline
ReLU-KAN   &  2  &  10  & 5  &  2.12E-01  &  1.38E-02  &  1.35E-01  &  8.67E-03 \\
ReLU-KAN   &  2  &  10  & 10  &  2.05E-01  &  2.63E-02  &  3.14E-01  &  2.00E-01 \\
ReLU-KAN   &  2  &  10  & 20  &  3.23E+00  &  4.75E-01   &  2.97E+00  &  1.43E-01 \\  \hline
Fast-KAN   &  2  &  10  & 5  &  2.35E-01  &  1.48E-02   &  1.37E-01  &  1.72E-02  \\
Fast-KAN   &  2  &  10  & 10  &  2.30E-01  &  1.15E-02  &  1.29E-01  &  1.02E-02 \\
Fast-KAN   &  2  &  10  & 20  &  2.24E-01  &  2.28E-02  &  1.43E-01  &  2.01E-02 \\  \hline
P1-KAN   &  2  &  10  & 5  &  2.15E-01  &  7.10E-03   &  1.21E-01  &  6.61E-03 \\
P1-KAN   &  2  &  10  & 10  &  {\bf 6.63E-02}  &  1.05E-02  &  1.01E-01  &  2.32E-02\\
P1-KAN   &  2  &  10  & 20  &  {\bf 2.27E-02}  &  3.56E-03 &     {\bf 1.93E-02}  &  3.90E-03\\  \hline
P1-KAN no adapt   &  2  &  10  & 5  &  2.25E-01  &  6.32E-03   &  1.22E-01  &  2.96E-03 \\
P1-KAN no adapt   &  2  &  10  & 10  &  1.70E-01  &  7.39E-03  &  1.09E-01  &  5.60E-03 \\
P1-KAN no adapt   &  2  &  10  & 20  &  1.12E-01  &  1.38E-02   &  {\bf 9.18E-02}  &  9.35E-03\\ \hline
\end{tabular}
    \caption{\rd{Average MSE  and standard deviation  obtained from 10 runs for B function in dimension 4 and 5: P1-KAN with adaptation and high $P$ has the best accuracy.}}
    \label{tab:compP1SplineIrrg}
\end{table}

\section{Application to Hydraulic Valley Optimization}
Hydraulic valley optimization in countries with large valleys, such as France, Brazil, or Norway, is traditionally achieved using methods based on dynamic programming \cite{bellman1958dynamic}.
The objective to be maximized is the profit obtained by selling the electricity produced by the turbines of some interconnected reservoirs. These profits are therefore linked to some price scenarios, and the different reservoirs and turbines have to respect some constraints, either physical (size of the lake, availability of the turbines, etc.) or environmental (minimum flooding for fish, agriculture, minimum lake level for leisure activities, etc.), while facing the uncertainties linked to the inflows.\\
The optimization step is generally the week, i.e., for each hour of a week, anticipative controls are computed, assuming that all uncertainties (inflows, prices) are known at the beginning of the week. Using dynamic programming type methods, Bellman values are computed at the beginning of each week, maximizing the gain in expectation.
\rd{}
Here, we test neural networks by comparing the results obtained with those obtained using traditional methods in production.\\

We suppose here that decisions are taken every week to turbine water using production units.
During week $i \in [1,52]$, time is discretized with three time steps per day. At the beginning of the week, turbining decisions are taken for each unit for each time step $t_{i,j}$, for $j=0, \ldots,20$ according to a flow equation given in Section \ref{sec:flowEq}. We take the convention $t_i= t_{i,0}$.
The energy generated and the function to maximize are given in Section \ref{sec:opt}.

\subsection{The Flow Equation During a Week}
\label{sec:flowEq}
We suppose that the valley is composed of $\hat{N}$ reservoirs and $\tilde{N}$ is the number of production units operating. 
\rd{A production unit $m$ of a reservoir $l$ is numbered $U^{l,m}$ for $m =1, \ldots,\tilde{N}(l)$.} Therefore $\tilde{N}(l)$ is the number of production units of the reservoir $l$. \\
A reservoir $r$ receives water:
\begin{itemize}
    \item due to inflow $I^r$ (rain, uncontrolled rivers, etc.),
    \item from one of the $N^{u}(r)$ upstream controlled reservoirs: either because an upstream reservoir is overflowing, or because a production unit number $U^{l,m}$ associated with an upstream reservoir $l$ is operating. $w(r,k)$ for $k=1, \ldots, N^{u}(r)$ is the number of the $k^{th}$ reservoir above reservoir $r$.
\end{itemize}
A reservoir \rd{$l$} releases water into one of its downstream reservoirs:
\begin{itemize}
    \item when it is overflowing,
    \item when one of its production units $U^{l,m}$ is activated to generate power, releasing water to another reservoir.
\end{itemize}
The flow equation for the volume $ \tilde V_{i,j+1}^r$ of a reservoir $r$ is given in week $i$ at time $t_{i,j+1}$ from the volume $V_{i,j}^r$  at the previous date without taking into account its overflow  by:
\begin{align}
    \tilde V^r_{i,j+1} = V^r_{i,j} + I^r_{i,j} + \sum_{k=1}^{N^u(r)} \big( O^{w(r,k)}_{i,j} + \sum_{m=1}^{\tilde N(w(r,k))} T^{U^{w(r,k),m}}_{i,j} \Delta_{i,j}\big) - \sum_{m=1}^{\tilde N(r)} T^{U^{r,m}}_{i,j} \Delta_{i,j} 
    \label{eq:flowEquation}
\end{align}
where:
\begin{itemize}
    \item $\Delta_{i,j} = t_{i,j+1} - t_{i,j}$ is the time step,
    \item $O^k_{i,j}$ is the volume of water overflowed by the upstream lake $k$,
    \item $T^k_{i,j}$ is the water turbined from the production unit $k$ per time unit.
\end{itemize}
The reservoir $r$ has a primary constraint that cannot be violated: $0 \le V_{i,j}^r \le \bar{V}^r$ for all $i, j$ and the volume of water overflowed satisfies $O^r_{i,j}= (\tilde V^r_{i,j+1} - \bar{V}^r)^{+}$, so that
\begin{align*}
    V^r_{i,j+1} =  \min( \tilde V^r_{i,j+1},\bar{V}^r).
\end{align*}
Environmental constraints are added but can be violated:
\begin{itemize}
\item $\underline{W}_{i,j}^r \le V_{i,j}^r \le \bar{W}_{i,j}^r$, for all $i,j$ where $\underline{W}^r$ and $\bar{W}^r$ are given,
\item $\underline{T}^k_{i,j} \le T^k_{i,j}$ where $\underline{T}^k$ is also given for each production unit $k$.
\end{itemize}
Moreover, the maximum turbining for a production unit $k$ associated with a reservoir $r$ is a function of the water level in  the reservoir $r$, giving the constraints that cannot be violated:
\begin{align}
\label{eq:TP}
    T^k_{i,j} \le \bar{T}^k(V_{i,j}^r),
\end{align}
where $\bar{T}^k$ are given functions associated with the production unit.
\subsection{Expected Gain to Maximize}
\label{sec:opt}
We denote $T= (T^k_{i,j})_{i=1,\ldots 52, j=0 \ldots 20, k=1, \ldots, \tilde N}$ where the $T^k_{i,j}$ are functions of the available information at the beginning of week $i$, assuming that the $(I^r_{i,j})_{j=0,20, r=1, \ldots, \hat N}$, $(S_{t_{i,j}})_{j=0,20}$ are known. Each production unit $k$ associated with lake $r$ transforms the volume turbined $T$ per unit time into power by a function $\phi^k(T,V^r)$ which is an increasing function of the level of the reservoir.\\ 
The expected gain is given by selling the electricity at a price $S_t$ and the function to maximize is given over 52 weeks by:
\begin{align}
\label{eq:obj}
    J(T)= G(T) - \epsilon \xi(T),
\end{align}
with the gain given as
\begin{align}
\label{eq:Gain}
    G(T)= \sum_{r=1}^{\hat N} \sum_{k=1}^{ \tilde N(r)}  \sum_{i=1}^{52} \sum_{j=0}^{20} \mathbb{E}[ S_{t_{i,j}} \phi^{U^{r,k}}(T^{U^{r,k}}_{i,j},V^r_{i,j}) \Delta_{i,j}],  
\end{align}
the volume violated by
\begin{align}
\label{eq:viol}
   \xi(T)= \sum_{r=1}^{\hat N} \sum_{i=1}^{52}\sum_{j=0}^{20}
   \big(\mathbb{E}[(\underline{W}^r_{i,j} -V^r_{i,j})^{+}+ ( V^r_{i,j} -\bar{W}^r_{i,j} )^+] + 
    \Delta_{i,j} \sum_{k=1}^{ \tilde N(r)} \mathbb{E}[  (\underline{T}^{U^{r,k}}_{i,j} - T^{U^{r,k}}_{i,j})^+]  \big)
\end{align}
where $\epsilon$ is a penalty factor.
\rd{This penalty factor is necessary because there are inflow scenarios in which it is impossible to satisfy all constraints. Therefore, a trade‑off is required, allowing some non‑physical constraints to be violated occasionally. Moreover, in production, managers do not want to avoid such violations at any cost.
Environmental constraints do not incur a direct economic cost. However, failing to comply with them may expose the company to legal action and, hypothetically, to financial penalties if it cannot demonstrate that no alternative course of action was possible. Besides, the company’s image could be tarnished.}

\subsection{Parametrization of the Neural Networks}
We use 52 neural networks $\phi$ parametrized by $(\theta_i)_{i=1,\ldots, 52}$ for the problem (one per week).
As input for the network $i$ corresponding to week $i$, we take:
\begin{itemize}
    \item the volume of each lake $V=(V^k_{i,0})_{k=1,\ldots, \hat{N}}$ at the beginning of the week (at date $t_{i,0}$),
    \item an inflow scenario for week $i$ for all the reservoirs: $I=(I^k_{i,j})_{j=0,\ldots,20, k=1, \ldots, \hat{N}}$,
    \item a price scenario for week $i$: $S =(S_{t_{i,j}})_{j=0, \ldots, 20}$.
\end{itemize}
Using a sigmoid function as the activation function for the output layer, the neural network outputs $\phi^{\theta_i}(V,I,S) \in [0,1]^{\tilde{N} \times 20}$.
As in \cite{warin2023reservoir}, the turbining control of each production unit at each date $t_{i,j}$, $j=0, \ldots, 20$ is obtained from its  maximum allowed $(\bar{T}_{i,j}^k(V^r_{i,j}))_{j=0, \ldots, 20, k=1, \ldots, \tilde{N}}$ by
\begin{align}
   \tilde{T}_{i,j}^{\theta_i,k}= \bar{T}_{i,j}^k(V^r_{i,j}) \phi^{\theta_i}(V,I,S)_{k,j}, \text{ for } i=1, \ldots, 52, \quad j= 0, \ldots, 20, \quad k=1, \ldots, \tilde{N}.
\end{align}
Setting $\tilde{T}^{\theta}= (\tilde{T}_{i,j}^{\theta_i,k})_{i=1, \ldots, 52, j=0, \ldots, 20, k=1, \ldots, \tilde{N}}$, $\theta = (\theta_i)_{i=1,\ldots,52}$, we are left to calculate
\begin{align}
\label{eq:objNN}
    \theta^*= \argmax_{\theta} G( \tilde{T}^\theta) - \epsilon \xi( \tilde{T}^\theta).
\end{align}

Classically, we solve \eqref{eq:objNN} using the Adam stochastic gradient algorithm with a learning rate equal to $10^{-3}$. The batch size is taken equal to 2000.
The number of inflow scenarios is limited (equal to 42) and a scenario generator allows us to build as many scenarios as we wish to avoid over-fitting.
\rd{126 price scenarios are generated by a software calculating marginal prices associated with the global electrical network,   and a price generator can then produce as many scenarios as needed from these initial scenarios to feed the stochastic gradient.}
\rd{The adapted P1-KAN network is employed, and its results are compared with those obtained using MLPs and other KANs.}

\subsection{The Test Case and the Results}
\begin{figure}[H]
    \centering
 \includegraphics[width=0.4\textwidth]{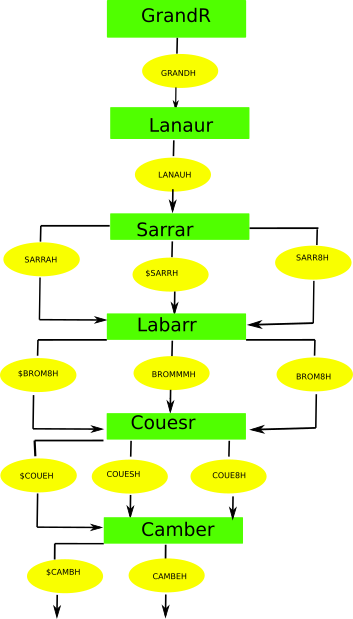}
 \caption{\label{fig:structVal} Structure of the valley}
\end{figure}

Figure \ref{fig:structVal} shows the structure of the valley: green squares represent reservoirs and yellow circles represent production units.
\rd{A reservoir can have a different number of production units emptying itself and fulling the reservoir below.
In equation~\eqref{eq:flowEquation}, for example for $r=\mathrm{Labarr}$, we have
\[
\begin{aligned}
N^u(\mathrm{Labarr}) &= 1, \\
w(\mathrm{Labarr},1) &= \mathrm{Sarrar}, \\
\tilde N\bigl(w(\mathrm{Labarr},1)\bigr) &= 3, \\
\bigl(U^{w(\mathrm{Labarr},1),k}\bigr)_{k=1,\ldots,3}
  &= (\mathrm{Sarrah},\, \text{\$}\mathrm{Sarrh},\, \mathrm{Sarr8h}), \\
\tilde N(\mathrm{Labarr}) &= 3, \\
\bigl(U^{\mathrm{Labarr},k}\bigr)_{k=1,\ldots,3}
  &= (\text{\$}\mathrm{Brom8h},\, \mathrm{Brommmh},\, \mathrm{Brom8h}).
\end{aligned}
\]
Here, $\hat N= 6$ gives for the $52$ weeks a state dimension equal to $52 \times \big( 6 \times (1+21) + 21)= 7956$ and as the number of unit production is equal to $13$, the dimension of the output is $52 \times 13\times 21 = 13923$.}

\begin{figure}[H]
  \begin{minipage}[b]{0.49\linewidth}
  \centering
 \includegraphics[width=\textwidth]{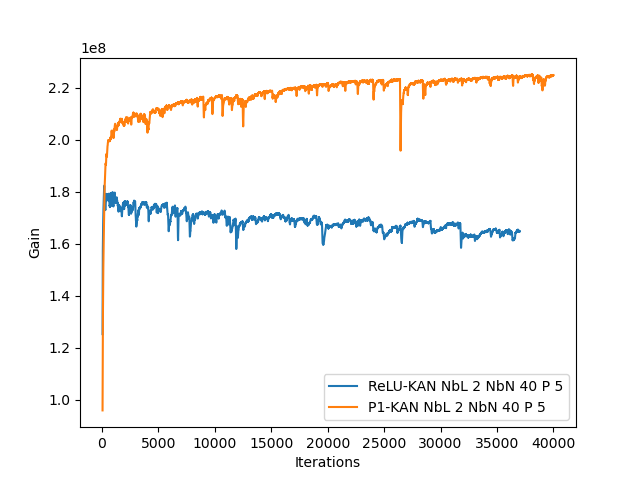}
 \caption{\rd{Comparison of the P1-KAN and the ReLU-KAN at the beginning of the valley  optimization ($\epsilon= 5\times10^{-3}$): slow divergence of the ReLU-KAN.\label{fig:valleyReLU} }}
 \end{minipage}
 \begin{minipage}[b]{0.49\linewidth}
  \centering
 \includegraphics[width=\textwidth]{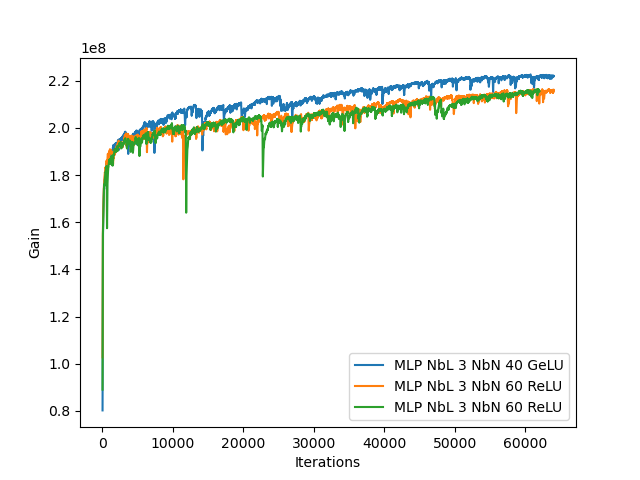}
 \caption{\rd{Comparison of the MLP activation function ($\epsilon= 5\times10^{-3}$): ReLU versus GeLU. GeLU converges slightly faster. \label{fig:valleyGeLU} }}
 \end{minipage}
\end{figure}

Due to its characteristics, this valley is difficult to optimize using classical methods based on dynamic programming.
The efficiency of the networks is measured by the expected gain from managing the valley. 
\rd{We were not able get results with the Fast-KAN network which was exploding with a few iterations.
The ReLU-KAN network begins to converge to the solution then slowly diverge as shown on figure \ref{fig:valleyReLU}.
For  the  MLP, the GeLU activation function \cite{hendrycks2016gaussian} is chosen for the comparison with KANs, as it yields slightly better results than the ReLU function  as shown on figure \ref{fig:valleyGeLU}.}
\begin{figure}[H]
    \begin{minipage}[b]{0.8\linewidth}
 \includegraphics[width=\textwidth]{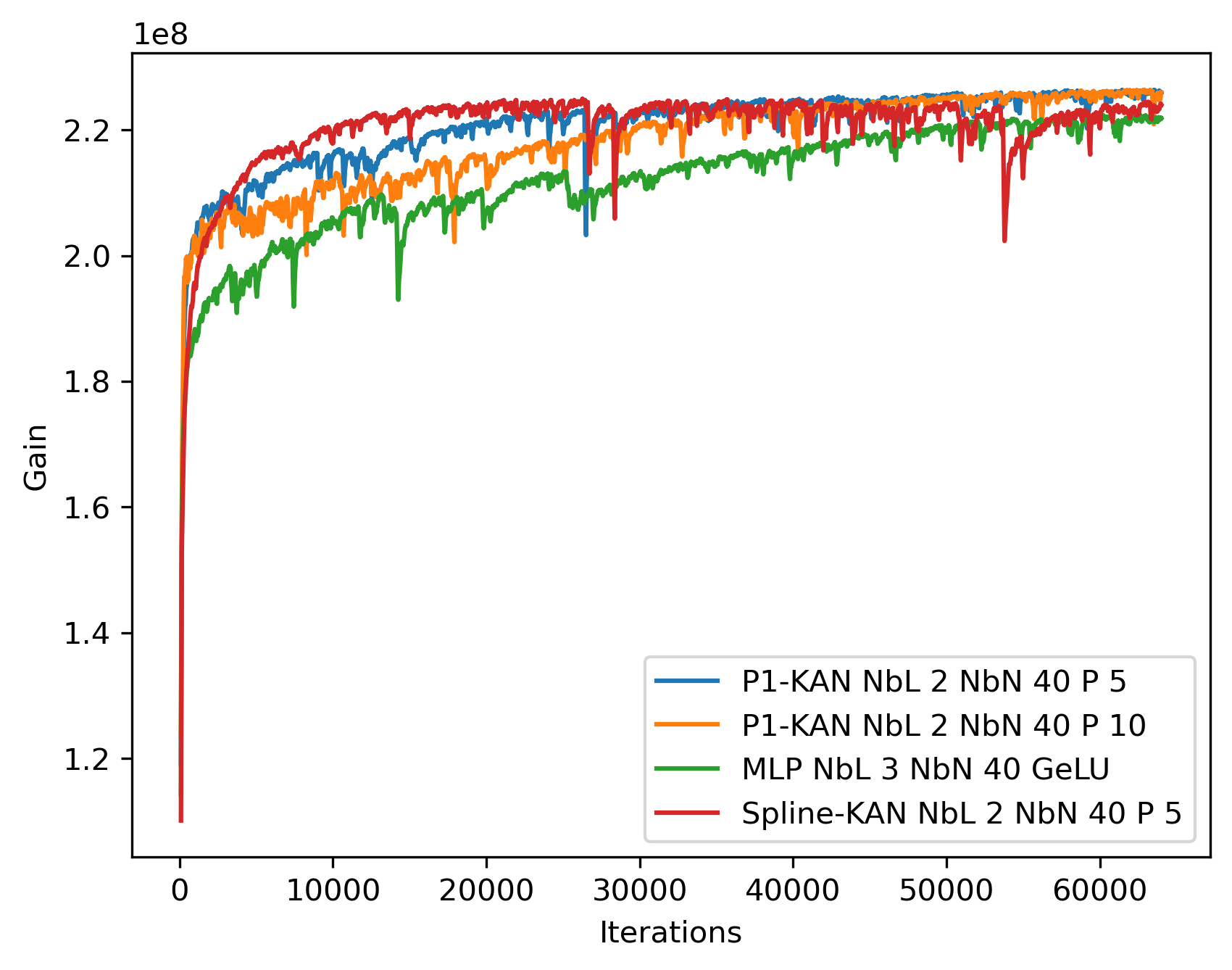}
 \caption*{$\epsilon= 5\times10^{-3}.$}
 \end{minipage}
   \begin{minipage}[b]{0.8\linewidth}
 \includegraphics[width=\textwidth]{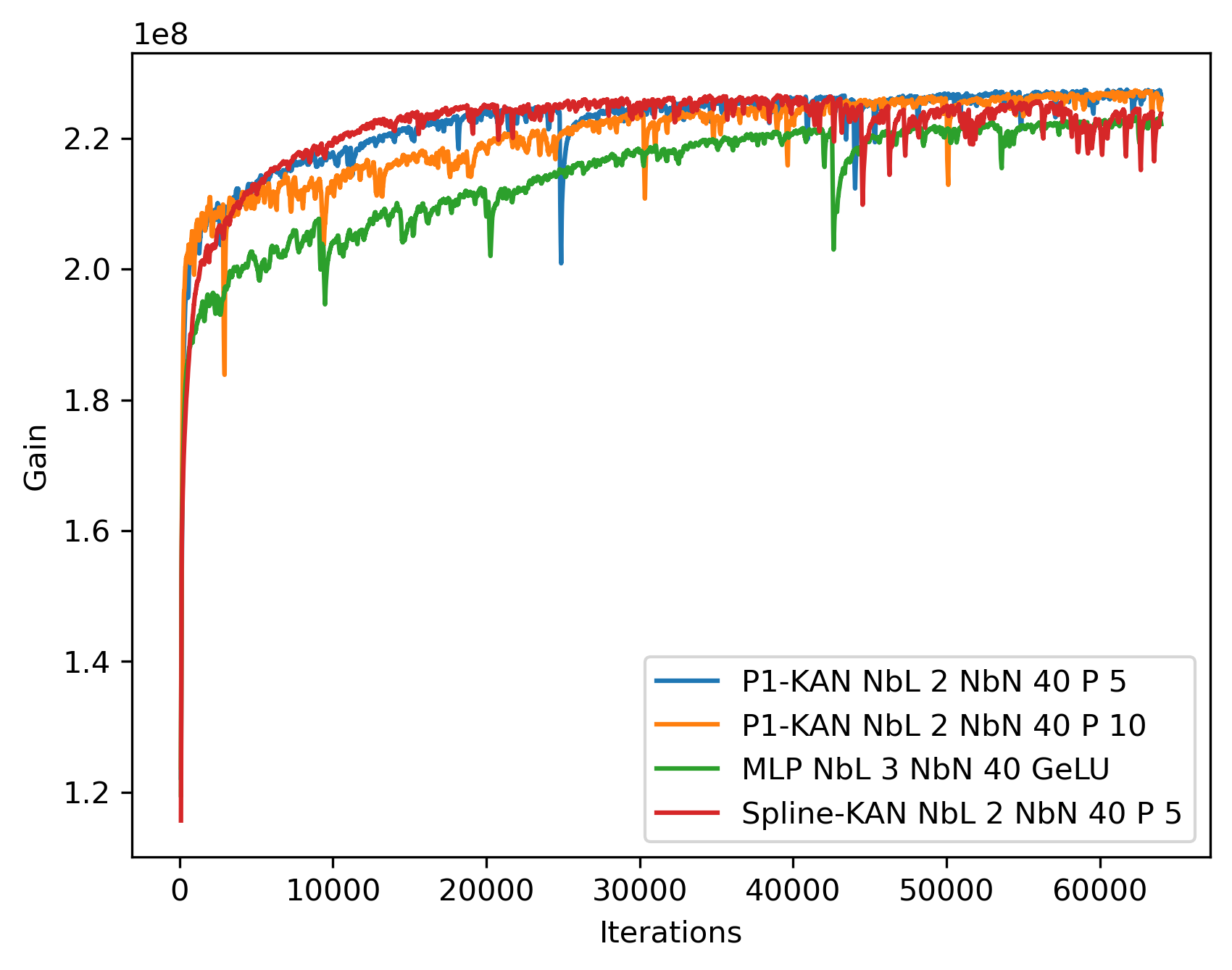}
 \caption*{$\epsilon= 2.5\times10^{-3}.$}
 \end{minipage}
 \caption{\label{fig:lossValleySim} Convergence of the expected gain in Euros (NbL: number of layers, NbN: number of neurons) \rd{: P1-KAN optimizes better using scenarios generated}.}
\end{figure}
\rd{Taking  $\epsilon= 5\times10^{-3}$ and $\epsilon= 2.5\times10^{-3}$}, with 65,000 gradient iterations and a rolling window of 50 successive results, the convergence plot (Figure \ref{fig:lossValleySim}) shows that the Spline-KAN and P1-KAN networks converge faster than the MLP. However, the P1-KAN network achieves a better optimization. The MLP appears capable of achieving a result superior to that of the Spline-KAN. Iteration time is similar for the different architectures on the NVIDIA H100 GPU (with a difference of less than 10 percent in computation time between networks for a given number of iterations). Most computation is unrelated to evaluating the controls by the networks, but rather, estimating the gain once the controls are calculated : \rd{one of the bottleneck seems to be the calculation of the power  function $\phi(T,V)$ appearing the gain function.} Global optimization of the 52 networks and 64,000 iterations takes around \rd{$10$} days per network.
\rd{The number of trainable parameters for the networks succeeding in optimizing the case is given in table \ref{tab:paramNum}.
 \begin{table}[H]
     \centering
     \begin{tabular}{|c|c|c|c|c|} \hline 
           Network &  nb layers & nb neurons & $P$ &  Number of parameters \\ \hline 
          MLP& 3 &  40 &  & 1116596 \\ 
          P1-KAN & 2 & 40 & 5 & 6143800\\
          P1-KAN & 2 & 40 & 10 & 11274640 \\
          Spline-KAN & 2 & 40 & 5 & 10129600 \\ \hline 
     \end{tabular}
     \caption{{\rd{Number of trainable parameters per model used in figure \ref{fig:lossValleySim}.}}}
     \label{tab:paramNum}
 \end{table}
 }

The software in production, based on dynamic programming (DP), uses linear programming resolution to solve transition problems during the week (see Chapter 6 \cite{gevret2018stochastic} for a description of the resolution using Bellman cuts).
It only sees the historical inflows (duplicated three times) and the 126 price scenarios. \rd{These set are used both in optimization and testing.} Penalties are tuned and settled specifically depending on the type of constraint violation. 
\rd{As DP faces the curse of dimensionality, two reservoirs, GrandR and Sarrar, are managed using Bellman values, while the others are handled using heuristics. The use of neural networks to optimize the valley should serve as a reference for tuning the deterministic software. Note that this type of optimization in production must be performed frequently throughout the year, and the price scenarios generated for each run can differ significantly in their distributions. Therefore, the neural networks would need to be retrained for each run.
}
In Table \ref{tab:refValley}, we present the gain function obtained and the volume of constraint violations with the scenarios used in production: \rd{ the calculation is achieved every 1000 iterations and the best result kept for each network.}
\rd{ Selecting $\epsilon = 5 \times 10^{-3}$ --- a value consistent with that employed in the production software ---, the P1-KAN enables us to obtain a performance gain of the same order as that delivered by the operational system, but with significantly fewer violations.
Selecting $\epsilon = 2.5 \times 10^{-3}$, the P1-KAN  allows us to obtain a higher gain than that achieved with DP for the same level of violations.}

\begin{table}[H]
\centering 
\begin{minipage}[b]{0.49\linewidth}
\centering 
\begin{tabular}{|c|c|c|} \hline 
Method & $G$ & $\xi$ \\ \hline 
DP & 209.6 & 1198 \\
P1-KAN P=5 & 208.2 & 661\\
P1-KAN P=10 & 209.0 & 670\\
Spline-KAN & 207.3 & 680 \\
MLP GeLU & 205.7 & 800 \\ \hline 
\end{tabular}
\caption*{$\epsilon = 5 \times 10^{-3}$.}
\end{minipage}
\begin{minipage}[b]{0.49\linewidth}
\centering 
\begin{tabular}{|c|c|c|} \hline 
Method & $G$ & $\xi$ \\ \hline 
DP & 209.6 & 1198 \\
P1-KAN P=5 &210.2& 1229 \\
P1-KAN P=10 & 210.9& 1209\\
Spline-KAN & 209.0 & 1183\\
MLP GeLU & 207.4 & 1601 \\ \hline 
\end{tabular}
\caption*{$\epsilon = 2.5 \times 10^{-3}$.}
\end{minipage}
\caption{Best expected gains $G$ in millions of Euros, and violations in $10^6 m^3$   : \rd{  P1-KAN  optimizes better using the 126 scenarios used in production. }} \label{tab:refValley} 
\end{table}

The approach using neural networks is far more robust as it is obtained using far more scenarios. \rd{} As the optimization by the software in production only takes two hours on fewer than 10 CPUs, the approach using machine learning, which takes \rd{10 days} to train, is not competitive but provides a reference and allows us to address some constraints that are impossible to handle with dynamic programming methods.

\section{Conclusion} 
\rd{
The P1‑KAN is a powerful network with proven convergence properties for function approximation. While it performs similarly to Spline‑KANs on smooth functions, it is particularly well suited to irregular ones. Experiments using different networks for stochastic optimization on a real‑world problem confirm that the adapted P1‑KAN achieves the best overall performance.
However, the broader use of this method, and of KANs in general, is constrained by its higher computational cost compared to MLPs. Although P1‑KANs remain relatively slow, emerging implementations may significantly reduce computation time, as demonstrated for Spline‑KANs. Consequently, just as Spline‑KANs increasingly replace MLPs for smooth data, P1‑KANs could serve a similar purpose for irregular or nonsmooth settings.
Extensions of this framework---such as the development of input--convex neural networks for optimal transport problems or efficient architectures for mean--field control problems based on graphons---have already been explored \cite{deschatre2025input,mekkaoui2026learning} and the present approach  can be used in many fields.
Moreover, as computational costs continue to decrease due to ongoing advances in hardware and methodological techniques, the industrial use case presented here could realistically transition to production within a few years. One of the key advantages of neural--network--based modeling lies in its flexibility to incorporate a wide range of constraints, as well as in the relative simplicity of its implementation when compared with deterministic methods.}
\subsection*{Conflict of interest }
None declared.
\subsection*{Data Availability}
Data available on request from the authors.

\newpage
\bibliographystyle{plain}
\bibliography{biblio}

@inproceedings{qiu2025relu,
  title={Relu-kan: New kolmogorov-arnold networks that only need matrix addition, dot multiplication, and relu},
  author={Qiu, Qi and Zhu, Tao and Gong, Helin and Chen, Liming and Ning, Huansheng},
  booktitle={2025 IEEE Smart World Congress (SWC)},
  pages={1686--1694},
  year={2025},
  organization={IEEE}
}

@inproceedings{so2025higher,
  title={Higher-order-ReLU-KANs (HRKANs) for solving physics-informed neural networks (PINNs) more accurately, robustly and faster},
  author={So, Chi Chiu and Yung, Siu Pang},
  booktitle={2025 IEEE World AI IoT Congress (AIIoT)},
  pages={1035--1042},
  year={2025},
  organization={IEEE}
}

@article{liu2024kan,
  title={Kan: Kolmogorov-arnold networks},
  author={Liu, Ziming and Wang, Yixuan and Vaidya, Sachin and Ruehle, Fabian and Halverson, James and Solja{\v{c}}i{\'c}, Marin and Hou, Thomas Y and Tegmark, Max},
  journal={arXiv preprint arXiv:2404.19756},
  year={2024}
}

@book{kolmogorov1961representation,
  title={On the representation of continuous functions of several variables by superpositions of continuous functions of a smaller number of variables},
  author={Kolmogorov, Andrei Nikolaevich},
  year={1961},
  publisher={American Mathematical Society}
}

@article{poggio2020theoretical,
  title={Theoretical issues in deep networks},
  author={Poggio, Tomaso and Banburski, Andrzej and Liao, Qianli},
  journal={Proceedings of the National Academy of Sciences},
  volume={117},
  number={48},
  pages={30039--30045},
  year={2020},
  publisher={National Acad Sciences}
}

@article{girosi1989representation,
  title={Representation properties of networks: Kolmogorov's theorem is irrelevant},
  author={Girosi, Federico and Poggio, Tomaso},
  journal={Neural Computation},
  volume={1},
  number={4},
  pages={465--469},
  year={1989},
  publisher={MIT Press}
}

@article{chaudhuri2021b,
  title={B-splines},
  author={Chaudhuri, Arindam},
  journal={arXiv preprint arXiv:2108.06617},
  year={2021}
}

@article{bozorgasl2024wav,
  title={Wav-kan: Wavelet kolmogorov-arnold networks},
  author={Bozorgasl, Zavareh and Chen, Hao},
  journal={arXiv preprint arXiv:2405.12832},
 year={2024}
}

@article{ss2024chebyshev,
  title={Chebyshev polynomial-based kolmogorov-arnold networks: An efficient architecture for nonlinear function approximation},
  author={SS, Sidharth},
  journal={arXiv preprint arXiv:2405.07200},
  year={2024}
}

@article{li2024kolmogorov,
  title={Kolmogorov-arnold networks are radial basis function networks},
  author={Li, Ziyao},
  journal={arXiv preprint arXiv:2405.06721},
  year={2024}
}

@article{yu2024kan,
  title={Kan or mlp: A fairer comparison},
  author={Yu, Runpeng and Yu, Weihao and Wang, Xinchao},
  journal={arXiv preprint arXiv:2407.16674},
  year={2024}
}

@article{knottenbelt2025coxkan,
  title={Coxkan: Kolmogorov-arnold networks for interpretable, high-performance survival analysis},
  author={Knottenbelt, William and McGough, William and Wray, Rebecca and Zhang, Woody Zhidong and Liu, Jiashuai and Machado, Ines Prata and Gao, Zeyu and Crispin-Ortuzar, Mireia},
  journal={Bioinformatics},
  volume={41},
  number={8},
  pages={btaf413},
  year={2025},
  publisher={Oxford University Press}
}

@article{cheon2024demonstrating,
  title={Demonstrating the efficacy of kolmogorov-arnold networks in vision tasks},
  author={Cheon, Minjong},
  journal={arXiv preprint arXiv:2406.14916},
  year={2024}
}

@article{yang2024kolmogorov,
  title={Kolmogorov-Arnold Transformer},
  author={Yang, Xingyi and Wang, Xinchao},
  journal={ICLR},
  year={2025}
}

@article{genet2024tkan,
  title={Tkan: Temporal kolmogorov-arnold networks},
  author={Genet, Remi and Inzirillo, Hugo},
  journal={arXiv preprint arXiv:2405.07344},
  year={2024}
}

@inproceedings{li2025u,
  title={U-kan makes strong backbone for medical image segmentation and generation},
  author={Li, Chenxin and Liu, Xinyu and Li, Wuyang and Wang, Cheng and Liu, Hengyu and Liu, Yifan and Chen, Zhen and Yuan, Yixuan},
  booktitle={Proceedings of the AAAI conference on artificial intelligence},
  volume={39},
  number={5},
  pages={4652--4660},
  year={2025}
}

@article{vaca2024kolmogorov,
  title={Kolmogorov-arnold networks (kans) for time series analysis},
  author={Vaca-Rubio, Cristian J and Blanco, Luis and Pereira, Roberto and Caus, M{\`a}rius},
  journal={arXiv preprint arXiv:2405.08790},
  year={2024}
}

@inproceedings{shen2025reduced,
  title={Reduced effectiveness of kolmogorov-arnold networks on functions with noise},
  author={Shen, Haoran and Zeng, Chen and Wang, Jiahui and Wang, Qiao},
  booktitle={ICASSP 2025-2025 IEEE International Conference on Acoustics, Speech and Signal Processing (ICASSP)},
  pages={1--5},
  year={2025},
  organization={IEEE}
}

@inproceedings{le2024exploring,
  title={Exploring the limitations of kolmogorov-arnold networks in classification: Insights to software training and hardware implementation},
  author={Le, Tran Xuan Hieu and Tran, Thi Diem and Pham, Hoai Luan and Le, Vu Trung Duong and Vu, Tuan Hai and Nguyen, Van Tinh and Nakashima, Yasuhiko and others},
  booktitle={2024 Twelfth International Symposium on Computing and Networking Workshops (CANDARW)},
  pages={110--116},
  year={2024},
  organization={IEEE}
}

@article{xu2024fourierkan,
  title={FourierKAN-GCF: Fourier Kolmogorov-Arnold Network--An Effective and Efficient Feature Transformation for Graph Collaborative Filtering},
  author={Xu, Jinfeng and Chen, Zheyu and Li, Jinze and Yang, Shuo and Wang, Wei and Hu, Xiping and Ngai, Edith C-H},
  journal={arXiv preprint arXiv:2406.01034},
  year={2024}
}

@article{bodner2024convolutional,
  title={Convolutional Kolmogorov-Arnold Networks},
  author={Bodner, Alexander Dylan and Tepsich, Antonio Santiago and Spolski, Jack Natan and Pourteau, Santiago},
  journal={arXiv preprint arXiv:2406.13155},
  year={2024}
}

@article{ta2024bsrbf,
  title={BSRBF-KAN: A combination of B-splines and Radial Basic Functions in Kolmogorov-Arnold Networks},
  author={Ta, Hoang-Thang},
  journal={arXiv preprint arXiv:2406.11173},
  year={2024}
}

@article{abueidda2025deepokan,
  title={Deepokan: Deep operator network based on kolmogorov arnold networks for mechanics problems},
  author={Abueidda, Diab W and Pantidis, Panos and Mobasher, Mostafa E},
  journal={Computer Methods in Applied Mechanics and Engineering},
  volume={436},
  pages={117699},
  year={2025},
  publisher={Elsevier}
}

@incollection{germain2021neural,
  title={Neural networks-based algorithms for stochastic control and PDEs in finance},
  author={Germain, Maximilien and Pham, Huy{\^e}n and Warin, Xavier and others},
  booktitle={Machine Learning and Data Sciences for Financial Markets: a guide to contemporary practices},
  publisher={Cambridge University Press},
  editor={Agostino Capponi and Charles-Albert Lehalle},
  year={2023}
}

@article{warin2023reservoir,
  title={Reservoir optimization and Machine Learning methods},
  author={Warin, Xavier},
  journal={EURO Journal on Computational Optimization},
  volume={11},
  pages={100068},
  year={2023},
  publisher={Elsevier}
}

@article{bellman1958dynamic,
  title={Dynamic programming and stochastic control processes},
  author={Bellman, Richard},
  journal={Information and control},
  volume={1},
  number={3},
  pages={228--239},
  year={1958},
  publisher={Elsevier}
}

@article{hendrycks2016gaussian,
  title={Gaussian error linear units (gelus)},
  author={Hendrycks, Dan and Gimpel, Kevin},
  journal={arXiv preprint arXiv:1606.08415},
  year={2016}
}

@article{hornik1990universal,
  title={Universal approximation of an unknown mapping and its derivatives using multilayer feedforward networks},
  author={Hornik, Kurt and Stinchcombe, Maxwell and White, Halbert},
  journal={Neural networks},
  volume={3},
  number={5},
  pages={551--560},
  year={1990},
  publisher={Elsevier}
}

@phdthesis{gevret2018stochastic,
  title={STochastic OPTimization library in C++},
  author={Gevret, Hugo and Langren{\'e}, Nicolas and Lelong, Jerome and Lobato, Rafael D and Ouillon, Thomas and Warin, Xavier and Maheshwari, Aditya},
  year={2018},
  school={EDF Lab}
}

@book{quarteroni2009numerical,
  title={Numerical models for differential problems},
  author={Quarteroni, Alfio},
  year={2009},
  publisher={Springer}
}

@article{xu2024kan,
  title={Are kan effective for identifying and tracking concept drift in time series?},
  author={Xu, Kunpeng and Chen, Lifei and Wang, Shengrui},
  journal={arXiv e-prints},
  pages={arXiv--2410},
  year={2024}
}

@article{xu2024kolmogorov,
  title={Kolmogorov-arnold networks for time series: Bridging predictive power and interpretability},
  author={Xu, Kunpeng and Chen, Lifei and Wang, Shengrui},
  journal={arXiv preprint arXiv:2406.02496},
  year={2024}
}

@article{lorentz1966approximation,
  title={Approximation of functions, athena series},
  author={Lorentz, GG},
  journal={Selected Topics in Mathematics},
  year={1966}
}

@article{fridman1967improvement,
  title={Improvement in the smoothness of functions in the Kilmogorov superposition theorem},
  author={Fridman, B},
  journal={Do\&l. Akad. Nauk. SSSR},
  volume={177},
  number={5},
  year={1967}
}

@inproceedings{henkin1964linear,
  title={Linear superpositions of continuously differentiable functions},
  author={Henkin, Gennadi Markovich},
  booktitle={Doklady Akademii Nauk},
  volume={157},
  pages={288--290},
  year={1964},
  organization={Russian Academy of Sciences}
}

@inproceedings{vitushkin1964proof,
  title={A proof of the existence of analytic functions of several variables not representable by linear superpositions of continuously differentiable functions of fewer variables},
  author={Vitushkin, Anatoliy Georgievich},
  booktitle={Doklady Akademii Nauk},
  volume={156},
  pages={1258--1261},
  year={1964},
  organization={Russian Academy of Sciences}
}

@article{song2025explicit,
  title={Explicit Construction of Approximate Kolmogorov-Arnold Superpositions with C2-Smoothness},
  author={Song, Lunji and Toscano, Juan Diego and Wang, Li-Lian},
  journal={arXiv preprint arXiv:2508.04392},
  year={2025}
}

@article{demb2021note,
  title={A note on computing with Kolmogorov Superpositions without iterations},
  author={Demb, Robert and Sprecher, David},
  journal={Neural Networks},
  volume={144},
  pages={438--442},
  year={2021},
  publisher={Elsevier}
}

@article{kratsios2025kolmogorov,
  title={Kolmogorov-Arnold Networks: Approximation and learning guarantees for functions and their derivatives},
  author={Kratsios, Anastasis and Furuya, Takashi},
  journal={arXiv preprint},
  year={2025}
}

@article{deschatre2025input,
  title={Input Convex Kolmogorov Arnold Networks},
  author={Deschatre, Thomas and Warin, Xavier},
  journal={arXiv preprint arXiv:2505.21208},
  year={2025}
}

@article{mekkaoui2026learning,
  title={Learning operators on labelled conditional distributions with applications to mean field control of non exchangeable systems},
  author={Mekkaoui, Samy and Pham, Huy{\^e}n and Warin, Xavier},
  journal={arXiv preprint arXiv:2603.21683},
  year={2026}
}
\end{document}